\pdfoutput=1
\documentclass[11pt]{article}

\usepackage[preprint]{acl}
\usepackage{times}
\usepackage{latexsym}
\usepackage{amsmath}
\usepackage[T1]{fontenc}
\usepackage[utf8]{inputenc}

\usepackage{microtype}

\usepackage{inconsolata}

\usepackage{times}
\usepackage{latexsym}

\usepackage{graphics}
\usepackage{graphicx}
\usepackage{color}
\usepackage{colortbl}
\usepackage{amsfonts}
\usepackage{amssymb}
\usepackage{amsmath}
\usepackage{nicefrac}
\usepackage{bbm}
\usepackage{booktabs}
\usepackage{microtype}
\usepackage{multirow}
\usepackage{multicol}
\usepackage{tabu}
\usepackage{amsthm}
\usepackage{footmisc}
\usepackage{soul}
\usepackage{arydshln}
\usepackage{wasysym}
\usepackage{array}
\usepackage{makecell}
\usepackage{scalerel}

\usepackage{algpseudocode}
\usepackage{algorithm}
\usepackage{hyperref}
\usepackage{xspace}
\usepackage{url}
\usepackage{utfsym}
\usepackage{dirtytalk}

\usepackage{float}
\usepackage{enumitem}
\usepackage{tikz}
\usepackage{tcolorbox}
\usepackage{siunitx}

\usepackage{subcaption} %  for subfigures environments 

\providetoggle{showcomments}
\settoggle{showcomments}{true}
\newlength\mylen
\iftoggle{showcomments}{%
    \newcommand{\resolved}[3][]{\ifstrequal{#1}{resolved}{\textcolor{green}{RESOLVED:}~\textbf{{\MakeUppercase #2:}}~{#3}}{\textbf{\MakeUppercase #2:}~#3}}

}

\newcommand{\prompttext}[1]{\textcolor{darkgray}{\texttt{\small{#1}}}}

\newtcolorbox[auto counter]{promptfloattop}[2][]{%
    float=!t,%
    blend before title=dash hang,%
    title={\textbf{Rationale~\thetcbcounter:} #2},%
    colback=black!5!white,%
    colframe=black!70!white,%
    #1}

\newtcolorbox[auto counter]{promptfloatbottom}[2][]{%
    float=!b,%
    blend before title=dash hang,%
    title={\textbf{Rationale~\thetcbcounter:} #2},%
    colback=black!5!white,%
    colframe=black!70!white,%
    #1}

\newcommand{\promptbottom}[3]{%
    \begin{promptfloatbottom}[label={#1}]{#2}
    {\vspace{3pt}\small \texttt{#3}\vspace{3pt}}
\end{promptfloatbottom}
}

\newtcolorbox[auto counter]{promptfloat}[2][]{%
    float=!h,%
    blend before title=dash hang,%
    title={\textbf{I\thetcbcounter:} #2},%
    colback=black!5!white,%
    colframe=black!70!white,%
    #1}

\newcommand{\prompt}[3]{%
    \begin{promptfloat}[label={#1}]{#2}
    {\vspace{8pt}\small \texttt{#3}\vspace{8pt}}
\end{promptfloat}
}

\newcolumntype{L}[1]{>{\let\newline\\\arraybackslash\hspace{0pt}}m{#1}}

\title{No Need for Explanations: LLMs can \textit{implicitly} learn from mistakes in-context}

\author{
 \textbf{Lisa Alazraki\textsuperscript{1}\thanks{Work done while at Cohere.}},
 \textbf{Maximilian Mozes\textsuperscript{2}},
 \textbf{Jon Ander Campos\textsuperscript{2}},
 \textbf{Tan Yi-Chern\textsuperscript{2}},
\\
 \textbf{Marek Rei\textsuperscript{1}},
 \textbf{Max Bartolo\textsuperscript{2}}
\\
 \textsuperscript{1}Imperial College London,
 \textsuperscript{2}Cohere
\\
\texttt{lisa.alazraki20@imperial.ac.uk}
}

\begin{document}
\maketitle

\begin{abstract}

Showing incorrect answers to Large Language Models (LLMs) is a popular strategy to improve their performance in reasoning-intensive tasks. It is widely assumed that, in order to be helpful, the incorrect answers must be accompanied by comprehensive rationales, explicitly detailing where the mistakes are and how to correct them.  However, in this work we present a counterintuitive finding: we observe that LLMs perform \textit{better} in math reasoning tasks when these rationales are eliminated from the context and models are left to infer on their own what makes an incorrect answer flawed. This approach also substantially outperforms chain-of-thought prompting in our evaluations. These results are consistent across LLMs of different sizes and varying reasoning abilities. To gain an understanding of \textit{why} LLMs learn from mistakes more effectively without explicit corrective rationales, we perform a thorough analysis, investigating changes in context length and answer diversity between different prompting strategies, and their effect on performance. We also examine evidence of overfitting to the in-context rationales when these are provided, and study the extent to which LLMs are able to autonomously infer high-quality corrective rationales given only incorrect answers as input. We find evidence that, while incorrect answers are more beneficial for LLM learning than additional diverse \textit{correct} answers, explicit corrective rationales over-constrain the model, thus limiting those benefits.
\end{abstract}

\vspace{2pt}

\section{Introduction}
\begin{figure}[!t]

\begin{subfigure}{\linewidth}
\includegraphics[width=\linewidth]{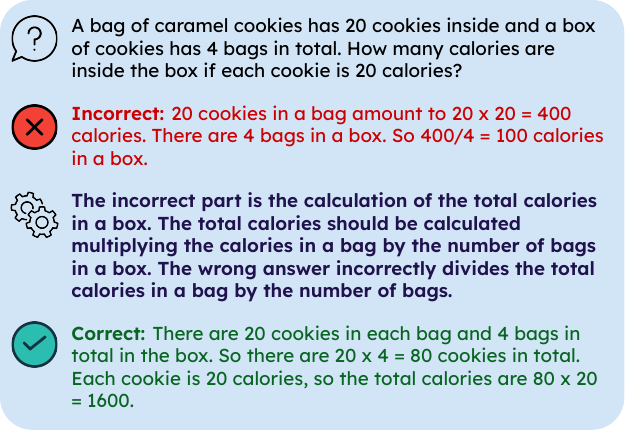}%\vspace{-4pt}
\caption{}
\label{fig:explicit}
\end{subfigure}
\begin{subfigure}{\linewidth}
\includegraphics[width=\linewidth]{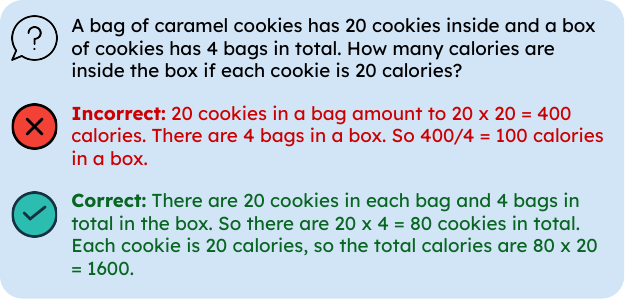}%\vspace{-4pt}
\caption{}
\label{fig:implicit}
\end{subfigure}
\caption{Learning examples for (a) explicit learning and (b) implicit learning. In explicit learning prompts, corrective feedback follows the incorrect answer (in red) and explains how to derive from it the correct answer (in green). For implicit learning, corrective feedback is discarded and the model is expected to infer the differences between the incorrect and the correct answer. 
}
\label{fig:prompts}
\end{figure}

Adding incorrect answers to the training examples of an LLM has become an established strategy to improve new generations, as these models are able to learn from their and others' mistakes \cite{MadaanSelfRefine, ShinnReflexion, an2023lema, paul-etal-2024-refiner}. Existing literature assumes that in order to learn from incorrect answers effectively, these must be accompanied by explicit corrective rationales locating each error and/or explaining how to rectify it. 

Previous work finds that when mistakes are shown to the model alongside this corrective feedback---whether in the training data \cite{an2023lema, paul-etal-2024-refiner} or in context \cite{MadaanSelfRefine, ShinnReflexion}---accuracy improves compared to fine-tuning or prompting with only valid question-answer pairs.

In this work, we investigate the ability of LLMs to learn from mistakes \textit{implicitly}---without the aid of corrective feedback---within an In-Context Learning (ICL) setting. 
We construct few-shot prompts for questions designed to probe reasoning abilities, alongside incorrect and correct chain-of-thought answers. We provide the incorrect answers without any rationales that would help contextualise them, and let the model autonomously infer the patterns that render an answer wrong or correct. We refer to this strategy as `prompting for \textit{implicit} learning'. We compare its performance against two baselines: (i) a prompt that includes the same correct and incorrect responses, but where the latter are accompanied by corrective rationales, as in \citet{an2023lema} (we refer to this strategy as `prompting for \textit{explicit} learning'), and (ii) a prompt only including the valid step-by-step answers to the same questions, as in the original chain-of-thought implementation \cite{10.5555/3600270.3602070} (we refer to this prompting strategy simply as `CoT'). We find not only that prompting for implicit learning outperforms vanilla CoT, but---remarkably---that it is also superior to the far more established strategy of prompting for explicit learning. To ensure the robustness of the results, we test all strategies extensively using seven LLMs from four distinct model families and four diverse tasks distributed across four established mathematical reasoning datasets.

Additionally, we carry out a thorough analysis to understand why implicit learning prompts outperform explicit ones. Firstly, we examine whether the greater context length that results from adding incorrect answers to the prompt contributes to the performance improvement---an investigation that was missing in prior literature on learning from mistakes. We insert additional valid question-answer pairs in the vanilla CoT setup, which matches the context length of implicit learning prompts and also provides the model with more diverse examples. We also experiment with showing two correct step-by-step answers for each question in the prompt. In both cases, we find that incorrect answers are more beneficial for LLM performance than additional correct answers. Secondly, to test whether LLMs can indeed infer the patterns that inform correct and incorrect answers implicitly, we have them generate rationales for new incorrect answers under each prompting strategy. We then perform a human evaluation study on the generated rationales. We observe that rationales produced with both implicit and explicit prompting are nearly identical in quality according to annotators. In contrast, rationales produced using only correct CoT answers are substantially lower quality. Lastly, we analyse to what extent LLMs overfit to in-context corrective feedback. We find relatively high similarity scores between the new rationales generated with explicit learning prompting and those provided in-context, indicating that the models are over-constrained by explicit learning prompts. This is further confirmed by our visual inspection of the generated rationales.

In summary, our experiments confirm the findings from previous literature that incorrect answers are particularly beneficial for LLM learning---indeed more so than additional correct answers---yet they also evidence that conditioning the model to in-context rationales, as is currently standard practice, may limit those benefits by adding unnecessary constraints. Since corrective rationales are normally produced by state-of-the-art close-sourced models \cite{an2023lema}, and are thus expensive to curate at scale, our findings in favour of rationale-free learning have real-world utility for researchers and developers.

Our main contributions are:
\begin{enumerate}
    \item We investigate \textit{implicit} learning from mistakes with LLMs and compare it to \textit{explicit} learning that uses both mistakes and rationales. To the best of our knowledge, no such investigation has been carried out before, and existing work relies heavily on explicit corrective feedback.
    \item We demonstrate that prompting for implicit learning outperforms explicit learning, as well as other strong ICL baselines. This indicates LLMs are well-suited for implicit learning.
    \item Our analytical experiments show that, while incorrect answers are beneficial for LLM learning, explicit corrective rationales limit those benefits by over-constraining the model.
    \item Our work brings into question the rationale behind rationales and offers a simple yet effective alternative.
\end{enumerate}

\section{Related Work}

Incorrect answers have been leveraged in prior work to improve LLM responses in challenging tasks. The existing literature can be largely categorised into three approaches: (i) \textit{Self-refinement}, where an LLM critiques its own erroneous generations, (ii) \textit{External feedback}, where corrective rationales are sourced from a distinct LLM, and (iii) \textit{Multi-agent debate}, where two or more models take turns at providing feedback for a previously generated response.

\paragraph{Self-refinement.} The self-refinement pipeline is well exemplified by \citet{MadaanSelfRefine}. They devise a framework where an LLM first answers a question, then generates feedback for that answer, and finally outputs a new answer based on the feedback. Note that the model is not fine-tuned and each step is elicited via prompting. The refinement process can be repeated multiple times until a stopping criterion is met, to iteratively improve the final answer. In \citet{10.5555/3692070.3694530}, not only does the LLM critique its own incorrect answer, but it also summarises what principles can be learned from it. \citet{KimNeurIPS23} and \citet{ShinnReflexion} adopt a similar strategy: the model executes a task and, based on the error signal received from the environment, outputs a self-reflection. This is then added to the LLM context in the next episode.

\paragraph{External feedback.} \citet{xu-etal-2024-pride} observe that self-refinement is inherently biased as LLMs tend to assess their own generations positively.  Hence, \citet{xu-etal-2024-llmrefine} propose a two-model system, where a base LLM answers a question and a fine-tuned model critiques the answer. Similarly, \citet{olausson2023self} find LLM self-critique to be biased in the context of code generation, and show that utilising a second, larger model as the critic allows for more substantial improvements in the task. \citet{tong-etal-2024-llms} feed a corpus of questions and incorrect answers to PaLM2 \cite{anil2023palm2technicalreport}, which outputs the type and reasons for each mistake. They show that fine-tuning Flan-T5 models \cite{chung2022scalinginstructionfinetunedlanguagemodels} on the resulting rationales improves their performance. Similarly, \citet{paul-etal-2024-refiner} use corrective feedback from a fine-tuned model as a signal to train a base LLM for producing better responses. \citet{an2023lema} extend the above approach by collecting LLM-generated incorrect answers and prompting GPT-4 \cite{openai2024gpt4technicalreport} to identify and correct the mistakes. They show that LLMs fine-tuned on this data achieve superior reasoning capabilities.

\paragraph{Multi-agent debate.} Since LLMs benefit from a single critic model, it is reasonable to assume that using multiple critics may achieve further improvements. Indeed, \citet{chen-etal-2024-reconcile} show that a round table of LLMs attains high accuracy in reasoning tasks. In their framework, each LLM produces an answer to a question, followed by a self-critique. Then, all models carry out a multi-turn discussion, revising their answers at each turn based on the responses and self-critiques of the other LLMs. \citet{du2023improvingfactualityreasoninglanguage} propose a similar framework where multiple instances of an LLM generate candidate answers to a math reasoning question. Each instance then critiques the output of the other models, and uses this to update its answer. \citet{khan2024debating} have two models generate different answers and debate their correctness, while the final choice is made by a third LLM witnessing the debate.

Lastly, related work that does not fall into the above categories is \citet{chia2023contrastivechainofthoughtprompting}'s contrastive CoT. Using an entity recognition model, they extract and randomly shuffle numbers and equations within a golden mathematical answer to obtain its incoherent counterpart. While this setup shares some 
similarities with implicit learning due to the absence of a rationale to accompany the incoherent answer, the latter is inherently different from our incorrect reasoning traces. Most saliently, their analysis is not concerned with what and how much information about previous mistakes is required to improve LLM reasoning. In contrast, our investigation stems from the observation that learning from mistakes with LLMs conventionally assumes the need for explicit, fine-grained corrective feedback. We seek to answer the previously unexplored question of whether this additional feedback is actually beneficial or even necessary.

\section{Prompt Construction}

Let $\, E = \big\|_{n=1}^N e_n$ be the text sequence resulting from concatenating $N$ in-context examples. In the typical few-shot CoT setting \cite{10.5555/3600270.3602070}, an individual example 
\begin{equation} 
e_n^{\mathrm{CoT}} = \big(q^{(n)}, a^{(n)}\big)
\label{eq:cot} 
\end{equation}
\noindent is defined by the question $q^{(n)}$ and the corresponding correct step-by-step answer $a^{(n)}$.
% as labelled in a given dataset.
This can be extended to 
\begin{equation} 
{e_n}^{\mathrm{explicit}} = \big(q^{(n)}, w^{(n)}, r^{(n)}, a^{(n)}\big)
\label{eq:explicit} 
\end{equation}
\noindent which additionally includes a wrong step-by-step answer $w^{(n)}$ and a rationale $r^{(n)}$ that explicitly identifies the errors in $w^{(n)}$ that need correcting to obtain $a^{(n)}$. This learning setup has been widely explored in prior literature \cite{MadaanSelfRefine, KimNeurIPS23, ShinnReflexion}. We additionally consider examples of the form 
\begin{equation} 
{e_n}^{\mathrm{implicit}} = \big(q^{(n)}, w^{(n)}, a^{(n)}\big)
\label{eq:implicit} 
\end{equation} 
\noindent where the explicit rationale $r^{(n)}$ is removed. Figure~\ref{fig:prompts} illustrates instances of (\ref{eq:explicit}) and (\ref{eq:implicit}).
In our experiments, we set $N = 8$ for all example types. We investigate how the example formulations in the set $\mathcal{E} = \lbrace E^{\mathrm{CoT}},\, E^{\mathrm{explicit}},\, E^{\mathrm{implicit}} \rbrace$ affect LLMs across different tasks: \textit{labelling} the correctness of an entire answer or an individual reasoning step, \textit{editing} an incorrect answer, or \textit{solving} a new question (we further elaborate on each task in Section \ref{sec:tasks}). We do not alter the example format by task, but we instead construct a task-specific prompt by appending an instruction, $I$,  to the examples. $I$ solely depends on the task and not on the type of examples preceding it. Hence, we have a set of task-specific instructions $\mathcal{I} = \lbrace I^{\mathrm{label_{ans}}},\, I^{\mathrm{label_{step}}}, \,I^{\mathrm{edit}}, \,I^{\mathrm{solve}} \rbrace$.

We experiment with all example types for all tasks. That is, we evaluate all prompts in the set $\mathcal{P} = \lbrace E\,  || \, I  \, \, | \, \,(E, I) \in \mathcal{E} \times \mathcal{I} \rbrace$. Prompts are shown in Appendix \ref{sec:prompts}.

\subsection{Generating Correct Answers}\label{sec:correct_gen}
All the examples in $\mathcal{E}$ include questions and their corresponding correct answers. While questions are provided by the training set, not all datasets contain CoT-style golden answers. In those cases, we generate them by prompting GPT-4 \cite{openai2024gpt4technicalreport} to provide answers in a zero-shot CoT fashion, and inspect both the reasoning trace correctness and final result.

\subsection{Generating Incorrect Answers}
The incorrect answers necessary to construct the exemplars in $E^{\mathrm{explicit}}$ and $E^{\mathrm{implicit}}$ are not present in most datasets. To obtain them, we prompt LLMs that are no longer state-of-the-art to generate answers for the training set questions. We use LLaMA 30B \cite{touvron2023llamaopenefficientfoundation}, Llama 2 7B \cite{touvron2023llama2openfoundation} and Llama 3 8B \cite{grattafiori2024llama3herdmodels}. The specific model choice depends on the dataset and its difficulty (refer to Appendix~\ref{sec:dataprep}). We use few-shot CoT prompting with all models. We gather answers that are marked as wrong by automated evaluation of the final numerical result. Having discarded empty and partial answers, we simply select the first $N$ incorrect answers in the set and pair them with the corresponding questions and their correct counterparts, obtained as detailed in Section \ref{sec:correct_gen}.

\begin{figure*}[!h]
    \centering
    \includegraphics[width=\textwidth]{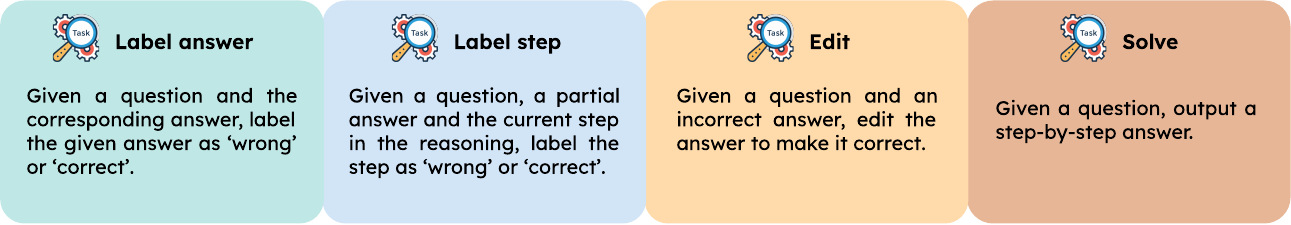}
    \caption{We evaluate LLM performance on four auxiliary tasks: (i) labelling an answer as wrong or correct, (ii) labelling an individual reasoning step, (iii) editing an incorrect answer to make it correct, and (iv)  solving a new question. In the labelling tasks, we instruct the model to output a rationale justifying its choice before generating the predicted label. This is motivated by previous work showing that this approach tends to produce more robust labels~\cite{trivedi2024selfrationalizationimprovesllmfinegrained, 10.5555/3666122.3668142}.}
    \label{fig:tasks}
\end{figure*}

\subsection{Generating Corrective Rationales}

We generate the corrective rationales in $E^{\mathrm{explicit}}$ following a strategy similar to that described in \citet{an2023lema}: we prompt GPT-4 in a few-shot fashion, showing it questions with incorrect and correct answers, as well as rationales. Given a new question and a pair of answers, we ask the model to identify the mistakes in the incorrect answer and explain how to correct them. We use the same few-shot examples as \citet{an2023lema}, slightly reformatted for our task. All rationales are carefully verified for correctness.
%(see Appendix \ref{sec:prompts}).

\section{Experiments}

\subsection{Models}
We study LLMs of different sizes: Command R\footnote{https://cohere.com/command} (35 billion parameters), Llama 3 70B Instruct \cite{grattafiori2024llama3herdmodels} (70 billion parameters), Command R+\footnotemark[\value{footnote}] (104 billion parameters), WizardLM \cite{xu2024wizardlm} (141 billion parameters). Note that for the Command models, we use both the original and the Refresh versions, as preliminary experiments showed significant differences in their output and results for math reasoning tasks. We also test Titan Text G1 Express\footnote{https://aws.amazon.com/bedrock/amazon-models}, whose exact number of parameters has not been publicly disclosed. We note, however, that this model is substantially less capable than the others in reasoning tasks, as evidenced by the lower scores in Table~\ref{tab:main_result}. Therefore, we consider seven LLMs in total. We employ a greedy sampling strategy with all models. LLMs are accessed via API; further details including the inference hyperparameters and model IDs are given in Appendix \ref{sec:models}.

\subsection{Datasets}

Our main focus is understanding whether LLMs learn implicitly in tasks that require complex reasoning. Contemporary work investigating LLM reasoning has primarily focused on math reasoning as an early and convenient proxy for complex reasoning ability evaluation \cite{ahn-etal-2024-large, paul-etal-2024-refiner, ruis2024proceduralknowledgepretrainingdrives, liu2025llmscapablestablereasoning}. Consistent with this approach, we test our method on several math reasoning benchmarks.
\vspace{-3pt}
\paragraph{GSM8K} includes grade-school-level arithmetic problems that require multiple reasoning steps to solve~\cite{cobbe2021trainingverifierssolvemath}. All problems in GSM8K can be tackled using basic arithmetic operations (addition, subtraction, multiplication, division).
\vspace{-3pt}
\paragraph{ASDiv} contains diverse problems of varying difficulty~\cite{miao-etal-2020-diverse}. In addition to arithmetic operations, questions can be solved using algebra, number theory, set operations and geometric formulas. They can also require pattern identification and unit conversion.
\vspace{-3pt}
\paragraph{AQuA} is a dataset of algebraic word problems from postgraduate admissions tests such as GRE and GMAT, as well as new questions of similar difficulty collected through crowd-sourcing~\cite{ling-etal-2017-program}. Note that while the original version of the dataset is multiple-choice, here we use a more challenging open-ended version.
\vspace{-3pt}
\paragraph{PRM800K}~\cite{lightman2024lets} is derived from the MATH dataset~\cite{2021_be83ab3e}, which contains challenging competition-level math problems. In PRM800K, model-generated answers to the questions in MATH are paired with human annotations providing a validation signal on intermediary reasoning steps.

These datasets cover a wide range of math domains and difficulty levels, each constituting a particular challenge. Furthermore, statistical analysis on GSM8K, ASDiv and AQuA has determined that these datasets are entirely out-of-domain with respect to one another \cite{hub_nature}, which makes this selection of evaluation datasets an appropriate test bed for our analysis.

\subsection{Tasks}\label{sec:tasks}

\begin{table*}[t!]
\centering
\renewcommand\arraystretch{1.15}
\setlength{\tabcolsep}{4.85pt}
\scalebox{0.68}{
\begin{tabular}{llccccccccccccc}
\hline
\toprule
{\multirow{2.2}{*}{\textbf{Model}}}
& {\multirow{2.2}{*}{\textbf{Strategy}}} 
& \multicolumn{3}{c}{\textbf{GSM8K}} 
& \multicolumn{3}{c}{\textbf{ASDiv}}
& \multicolumn{3}{c}{\textbf{AQuA}}
& \multicolumn{4}{c}{\textbf{PRM800K}} \\
\cmidrule(lr){3-5} \cmidrule(lr){6-8} \cmidrule(lr){9-11} \cmidrule(lr){12-15}
& & \textbf{\texorpdfstring{label\textsubscript{ans}}} & \textbf{edit} & \textbf{solve} & \textbf{\texorpdfstring{label\textsubscript{ans}}} & \textbf{edit} & \textbf{solve} & \textbf{\texorpdfstring{label\textsubscript{ans}}} & \textbf{edit} & \textbf{solve} & 
\textbf{\texorpdfstring{label\textsubscript{ans}}} & \textbf{\texorpdfstring{label\textsubscript{step}}} & \textbf{edit} & \textbf{solve} \\
\midrule
& CoT
& $83.8_{0.5}$ & $81.3_{0.3}$ & $91.8_{0.5}$
& $90.1_{0.3}$ & $82.7_{0.8}$ & $90.8_{0.3}$
& $\textbf{66.6}_{1.1}$ & $37.3_{0.4}$ & $55.8_{0.6}$
& $\textbf{31.7}_{0.6}$ & $49.6_{0.2}$ & $20.6_{1.3}$ & $43.9_{0.8}$ \\
\multirow{-1.4}{*}{\textit{\makecell[l]{Llama 3 70B \\ Instruct}}}
& Explicit
& $82.5_{0.6}$ & $84.2_{0.2}$ & $92.8_{0.3}$
& $90.0_{0.3}$ & $81.4_{0.9}$ & $\textbf{91.5}_{0.1}$
& $55.7_{1.1}$ & $34.0_{1.1}$ & $55.1_{1.0}$
& $19.0_{0.4}$ & $48.2_{0.4}$ & $21.8_{1.6}$ & $48.1_{0.4}$ \\
& Implicit
& $\textbf{84.0}_{0.7}$ & $\textbf{84.8}_{0.1}$ & $\textbf{93.3}_{0.4}$
& $\textbf{91.4}_{0.5}$ & $\textbf{84.9}_{0.7}$ & $91.1_{0.2}$
& $56.6_{0.2}$ & $\textbf{37.6}_{1.3}$ & $\textbf{56.4}_{0.4}$
& $19.2_{0.2}$ & $\textbf{50.0}_{0.6}$ & $\textbf{26.5}_{1.9}$ & $\textbf{48.4}_{0.6}$ \\
\cmidrule(lr){1-15}
& CoT
& $50.5_{0.8}$ & $17.2_{0.8}$ & $\textbf{63.1}_{0.5}$
& $53.4_{1.1}$ & $49.3_{1.2}$ & $\textbf{77.6}_{0.7}$
& $37.1_{0.8}$ & $7.9_{1.3}$ & $\textbf{21.9}_{1.1}$
& $21.4_{0.7}$ & $36.3_{0.4}$ & $4.7_{1.2}$ & $13.3_{1.0}$ \\
\multirow{-1}{*}{\textit{\makecell[l]{Command R}}}
& Explicit
& $57.0_{0.8}$ & $25.1_{1.3}$ & $56.7_{1.0}$
& $\textbf{64.1}_{1.2}$ & $48.0_{1.3}$ & $69.6_{1.0}$
& $34.2_{1.6}$ & $6.7_{1.1}$ & $17.8_{2.0}$
& $32.7_{0.6}$ & $39.0_{0.2}$ & $7.5_{0.8}$ & $13.0_{0.7}$ \\
& Implicit
& $\textbf{64.2}_{0.8}$ & $\textbf{31.1}_{1.2}$ & $60.5_{1.2}$
& $60.3_{0.7}$ & $\textbf{51.4}_{0.7}$ & $70.1_{0.6}$
& $\textbf{39.8}_{1.3}$ & $\textbf{11.2}_{1.2}$ & $19.1_{0.3}$
& $\textbf{56.0}_{0.9}$ & $\textbf{43.4}_{1.2}$ & $\textbf{8.8}_{0.4}$ & $\textbf{14.8}_{0.7}$ \\
\cmidrule(lr){1-15}
& CoT
& $65.8_{0.7}$ & $48.0_{0.3}$ & $69.7_{0.5}$
& $78.9_{0.6}$ & $61.9_{1.1}$ & $81.7_{0.6}$
& $43.8_{0.9}$ & $11.9_{1.2}$ & $32.0_{1.6}$
& $16.1_{1.1}$ & $35.8_{0.4}$ & $14.5_{1.1}$ & $\textbf{23.9}_{1.3}$ \\
\multirow{-1}{*}{\textit{\makecell[l]{Command R+}}}
& Explicit
& $64.3_{0.3}$ & $59.8_{0.6}$ & $76.0_{0.9}$
& $80.4_{0.3}$ & $69.3_{1.2}$ & $83.9_{0.4}$
& $46.5_{0.4}$ & $12.5_{1.4}$ & $31.1_{2.0}$
& $\textbf{59.7}_{0.8}$ & $38.8_{0.2}$ & $12.9_{1.0}$ & $18.1_{0.6}$ \\
& Implicit
& $\textbf{71.9}_{0.4}$ & $\textbf{62.0}_{0.8}$ & $\textbf{79.9}_{0.8}$
& $\textbf{82.6}_{0.2}$ & $\textbf{70.7}_{1.3}$ & $\textbf{85.3}_{0.3}$
& $\textbf{47.6}_{0.9}$ & $\textbf{16.8}_{1.1}$ & $\textbf{35.8}_{1.1}$
& $59.5_{1.2}$ & $\textbf{39.2}_{0.2}$ & $\textbf{16.6}_{0.8}$ & $21.1_{0.6}$ \\
\cmidrule(lr){1-15}
& CoT
& $55.5_{0.9}$ & $52.1_{0.6}$ & $78.9_{0.4}$
& $54.8_{0.3}$ & $64.8_{0.2}$ & $84.5_{0.6}$
& $47.5_{1.4}$ & $8.5_{1.0}$ & $35.3_{1.2}$
& $68.1_{0.7}$ & $39.3_{0.6}$ & $11.7_{0.7}$ & $30.6_{0.7}$ \\
\multirow{-1.4}{*}{\textit{\makecell[l]{Command R \\ Refresh}}}
& Explicit
& $48.7_{1.1}$ & $55.9_{0.4}$ & $75.9_{0.8}$ 
& $37.9_{0.6}$ & $69.2_{1.1}$ & $80.9_{0.6}$
& $42.4_{0.4}$ & $16.5_{0.5}$ & $39.1_{1.8}$
& $67.3_{0.9}$ & $\textbf{55.9}_{0.6}$ & $\textbf{13.1}_{0.8}$ & $30.8_{0.5}$ \\
& Implicit
& $\textbf{62.5}_{1.0}$ & $\textbf{57.4}_{0.7}$ & $\textbf{79.2}_{0.7}$
& $\textbf{70.4}_{1.1}$ & $\textbf{72.2}_{0.3}$ & $\textbf{84.8}_{0.5}$
& $\textbf{50.7}_{0.9}$ & $\textbf{16.6}_{0.6}$ & $\textbf{40.5}_{0.8}$
& $\textbf{71.1}_{0.9}$ & $53.7_{0.8}$ & $11.8_{1.1}$ & $\textbf{32.1}_{0.7}$ \\
\cmidrule(lr){1-15}
& CoT
& $46.9_{0.9}$ & $45.9_{0.7}$ & $75.6_{0.9}$
& $77.7_{0.1}$ & $78.8_{0.5}$ & $89.4_{0.4}$
& $61.0_{0.4}$ & $\textbf{23.5}_{0.7}$ & $44.5_{0.9}$
& $54.5_{1.2}$ & $\textbf{51.9}_{0.6}$ & $16.1_{1.3}$ & $\textbf{31.9}_{0.8}$ \\
\multirow{-1.4}{*}{\textit{\makecell[l]{Command R+ \\ Refresh}}}
& Explicit
& $40.3_{0.8}$ & $57.6_{1.3}$ & $82.0_{0.7}$
& $64.8_{0.2}$ & $76.1_{0.9}$ & $89.9_{0.3}$
& $53.6_{0.5}$ & $20.6_{1.7}$ & $43.2_{1.7}$
& $73.3_{0.4}$ & $51.7_{0.9}$ & $15.9_{1.3}$ & $26.8_{0.4}$ \\
& Implicit
& $\textbf{47.2}_{0.8}$ & $\textbf{62.8}_{0.9}$ & $\textbf{86.3}_{0.8}$
& $\textbf{79.6}_{0.2}$ & $\textbf{81.9}_{1.0}$ & $\textbf{90.4}_{0.4}$
& $\textbf{63.3}_{1.2}$ & $21.5_{1.1}$ & $\textbf{47.8}_{1.2}$
& $\textbf{73.9}_{0.8}$ & $47.8_{0.7}$ & $\textbf{18.7}_{1.1}$ & $29.7_{0.7}$ \\
\cmidrule(lr){1-15}
& CoT
& $53.4_{0.2}$ & $2.1_{0.3}$ & $29.7_{0.7}$
& $45.4_{0.6}$ & $\textbf{13.1}_{0.3}$ & $61.2_{0.4}$
& $\textbf{54.9}_{1.4}$ & $1.0_{0.2}$ & $7.9_{0.6}$
& $\textbf{67.2}_{0.4}$ & $46.4_{0.5}$ & $3.0_{0.3}$ & $5.4_{0.2}$ \\
\multirow{-1.4}{*}{\textit{\makecell[l]{Titan Text G1 \\ Express}}}
& Explicit
& $68.7_{0.3}$ & $2.8_{0.4}$ & $33.2_{0.5}$
& $53.0_{0.8}$ & $12.9_{0.3}$ & $61.1_{0.6}$
& $43.2_{1.3}$ & $1.2_{0.2}$ & $8.8_{0.7}$
& $35.5_{0.4}$ & $48.9_{0.4}$ & $3.0_{1.0}$ & $\textbf{7.6}_{0.4}$ \\
& Implicit
& $\textbf{69.3}_{0.5}$ & $\textbf{3.0}_{0.4}$ & $\textbf{34.7}_{0.9}$
& $\textbf{60.2}_{0.4}$ & $12.9_{0.1}$ & $\textbf{62.8}_{0.4}$
& $49.6_{0.8}$ & $\textbf{1.4}_{0.1}$ & $\textbf{11.0}_{0.2}$
& $45.7_{0.9}$ & $\textbf{49.7}_{1.0}$ & $\textbf{4.0}_{0.5}$ & $5.8_{0.5}$ \\
\cmidrule(lr){1-15}
& CoT
& $85.9_{0.4}$ & $72.4_{1.1}$ & $91.3_{0.6}$
& $93.8_{0.2}$ & $80.6_{0.3}$ & $90.9_{0.4}$
& $76.6_{0.2}$ & $41.7_{0.2}$ & $59.1_{0.7}$
& $68.4_{0.6}$ & $61.4_{0.4}$ & $\textbf{36.8}_{1.2}$ & $60.3_{0.9}$ \\
\multirow{-1}{*}{\textit{\makecell[l]{WizardLM}}}
& Explicit
& $85.4_{0.5}$ & $78.5_{1.3}$ & $\textbf{91.5}_{0.9}$
& $93.5_{0.3}$ & $82.4_{0.2}$ & $\textbf{91.4}_{0.1}$
& $77.3_{0.5}$ & $43.2_{0.4}$ & $58.0_{1.1}$
& $62.5_{0.4}$ & $60.2_{0.3}$ & $36.5_{1.5}$ & $60.8_{0.7}$ \\
& Implicit
& $\textbf{86.9}_{0.5}$ & $\textbf{82.3}_{0.9}$ & $\textbf{91.5}_{0.7}$
& $\textbf{95.0}_{0.2}$ & $\textbf{82.6}_{0.2}$ & $90.6_{0.3}$
& $\textbf{79.5}_{0.1}$ & $\textbf{44.1}_{0.5}$ & $\textbf{62.1}_{0.6}$
& $\textbf{69.0}_{0.4}$ & $\textbf{62.6}_{0.5}$ & $35.7_{1.3}$ & $\textbf{61.1}_{0.6}$ \\

\bottomrule
\hline
\end{tabular}
}
\caption{Results of CoT prompting, \textit{explicit} learning prompting, and \textit{implicit} learning prompting for different LLMs on four math reasoning benchmarks. We use the benchmarks for the following tasks: (i) labelling an answer as wrong or correct (label\textsubscript{ans}), (ii) labelling a single reasoning step as wrong or correct (label\textsubscript{step}), (iii) editing an incorrect answer (edit), and (iv) solving a question (solve). We report the accuracy of the final numerical result for all tasks except the two labelling tasks, where we report the weighted F1-score of the binary label. Due to small variations likely resulting from dynamic batching in the APIs, we report results averaged over five runs. Confidence intervals are shown in Appendix~\ref{sec:statistics}.
}\label{tab:main_result}
\end{table*}

In addition to evaluating on diverse math reasoning datasets, we consider auxiliary tasks that can be carried out within those datasets. We illustrate them below and in Figure~\ref{fig:tasks}.
\paragraph{Labelling an answer.} In this task, we have the model assign a binary label to a CoT-style answer, identifying whether it is correct or not, given the question. Previous work has found that LLM-assigned labels are more robust when they are accompanied by a model-generated rationale \cite{trivedi2024selfrationalizationimprovesllmfinegrained, 10.5555/3666122.3668142}. Hence, we require LLMs to first output a rationale explaining their choice, followed by the label. Performance in the binary labelling tasks is measured by the macro-averaged F1-score, weighted by support to account for label imbalance. The answers to be labelled are generated by running Llama 2 7B and Llama 3 8B on the test set of each dataset (refer to Appendix \ref{sec:dataprep} for details).\vspace{-2pt}
\paragraph{Labelling a reasoning step.} We leverage the step-wise reasoning annotations in PRM800K to have models score the correctness of a single reasoning step given the question and any previous context. Similar to the above setting, the LLM outputs a rationale followed by a binary label (`correct' or `incorrect'). As the other datasets do not contain step-wise annotations, we perform this task only on PRM800K.\vspace{-2pt}
\paragraph{Editing an incorrect answer.} We show the model a question and a corresponding incorrect answer, then ask it to output a new, edited answer that leads to the correct solution. Performance is measured by computing the accuracy of the numerical solution. For this task, we use the incorrect portion of the pre-generated answers obtained by running Llama 2 7B and Llama 3 8B on the test sets.\vspace{-2pt}
\paragraph{Solving a math question.} We show the model a test set question and ask it to output the solution. As in the previous task, we compute the accuracy of the final numerical solution.

To encourage the models to output responses \textit{conditioned} on the context, as opposed to text that merely mimics the format of the examples in it, we append the task-specific instruction after the examples. We further aid generalisation by prepending the text \textit{`Now apply what you have learned'} to the instruction. \citet{mao-etal-2024-prompt} show that the position of the instruction within a few-shot prompt affects the model's behaviour and performance. On the other hand, the model may still be inclined to generate responses in the format of the examples (e.g., when tasked with editing an answer, having observed examples that contain corrective rationales, the model may output a rationale before the corrected answer). To account for this possibility without unnecessarily penalising any particular prompting strategy, we provide a large generation window of 4096 tokens.

%%%%%%%%%%%%%%%%%

\subsection{Results}

\begin{figure}[!b]
\begin{subfigure}{.5\linewidth}
\includegraphics[width=\linewidth]{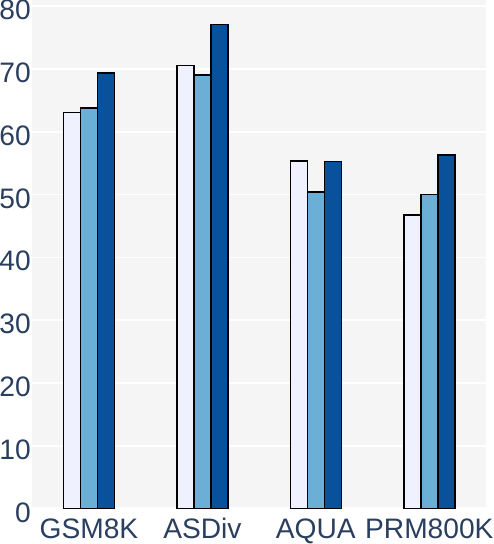}\vspace{-3pt}
\caption{}
\label{fig:label}
\end{subfigure}\begin{subfigure}{.5\linewidth}
\includegraphics[width=\linewidth]{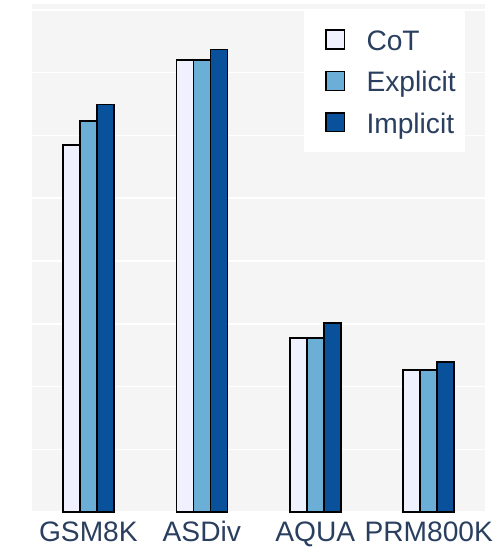}\vspace{-3pt}
\caption{}
\label{fig:edit_solve}
\end{subfigure}
\caption{Scores per dataset of CoT, explicit and implicit prompting for (a) the weighted F1-score of the labelling task, and (b) the averaged accuracy across the editing and solving tasks. Scores are averaged across all LLMs.
}
\label{fig:results_per_dataset}
\end{figure}

We find that CoT and prompting with \textit{explicit} rationales have similar overall performance on the answer labelling task and when solving new questions, while the latter outperforms CoT when labelling reasoning steps ($+$3.2\%, averaged across all models and all datasets) and editing an incorrect answer ($+$2.1\% avg.). This advantage is aligned with previous findings that LLMs benefit from observing incorrect answers and corrective feedback in their context. On the other hand, prompting for \textit{implicit} learning achieves the highest overall performance, as evidenced in Table \ref{tab:main_result}. When considering all combinations of model, dataset and task, implicit learning outperforms CoT in 85\% of cases. It also outperforms explicit learning in 88\% of cases. In nearly half of these, the advantage of implicit over explicit learning is substantial---well above 3\%. This advantage is present even in tasks where, intuitively, we would expect in-context rationales to be particularly helpful, for example when editing an incorrect answer to make it correct. In fact, implicit learning gives the largest accuracy boost in the editing task: $+4.4\%$ over CoT and $+2.2\%$ over explicit learning, averaged across all models and datasets. On the solving task, its accuracy increases by $1.6$ and $1.9$ percentage points, respectively. Labelling answers also benefits from implicit learning prompts, with averaged F1-scores $5.6\%$ above CoT and $6.2\%$ above explicit learning.
Finally, looking at the individual datasets, implicit learning gains the most on GSM8K, where it outperforms both explicit learning and CoT in over $90\%$ of cases across all models and tasks. This proportion is $76\%$ on ASDiv, $81\%$ on AQuA and $64\%$ on PRM800K. Note that the questions in GSM8K and ASDiv have a lower level of difficulty than those in AQuA and PRM800K, as evidenced by the performance differences across all LLMs. Generally, we observe that prompting for implicit learning improves performance across varying levels of difficulty, as shown in Figure~\ref{fig:results_per_dataset}. In the labelling task (Figure~\ref{fig:label}), implicit learning gives the most substantial performance gains on ASDiv and PRM800K. When editing an incorrect answer and solving a new question (Figure~\ref{fig:edit_solve}), on the other hand, it is GSM8K and AQuA that benefit the most from this strategy.

\section{Analysis}

To understand \textit{why} implicit learning leads to the improved performance observed above, we carry out a thorough analysis. We investigate context length and answer diversity, and draw insights from new rationales generated under each prompting strategy.

\subsection{Effect of Context Length and Diversity}\label{sec:context_length}

Adding incorrect answers to a prompt introduces additional tokens into the context. As a result, there is a mismatch between the context length of CoT and that of implicit learning. Since an extended context length can in itself be responsible for improved performance, we investigate whether the additional tokens may be driving the improvement, rather than the presence of incorrect answers. We thus compare implicit learning with two distinct extended-context baselines: a few-shot CoT prompt containing additional valid question-answer pairs (we refer to this setup as CoT+), and one where two correct step-by-step answers are shown for each question (we refer to this as CoT-2).\vspace{-2pt}

\begin{figure}[!t]
\includegraphics[width=\linewidth]{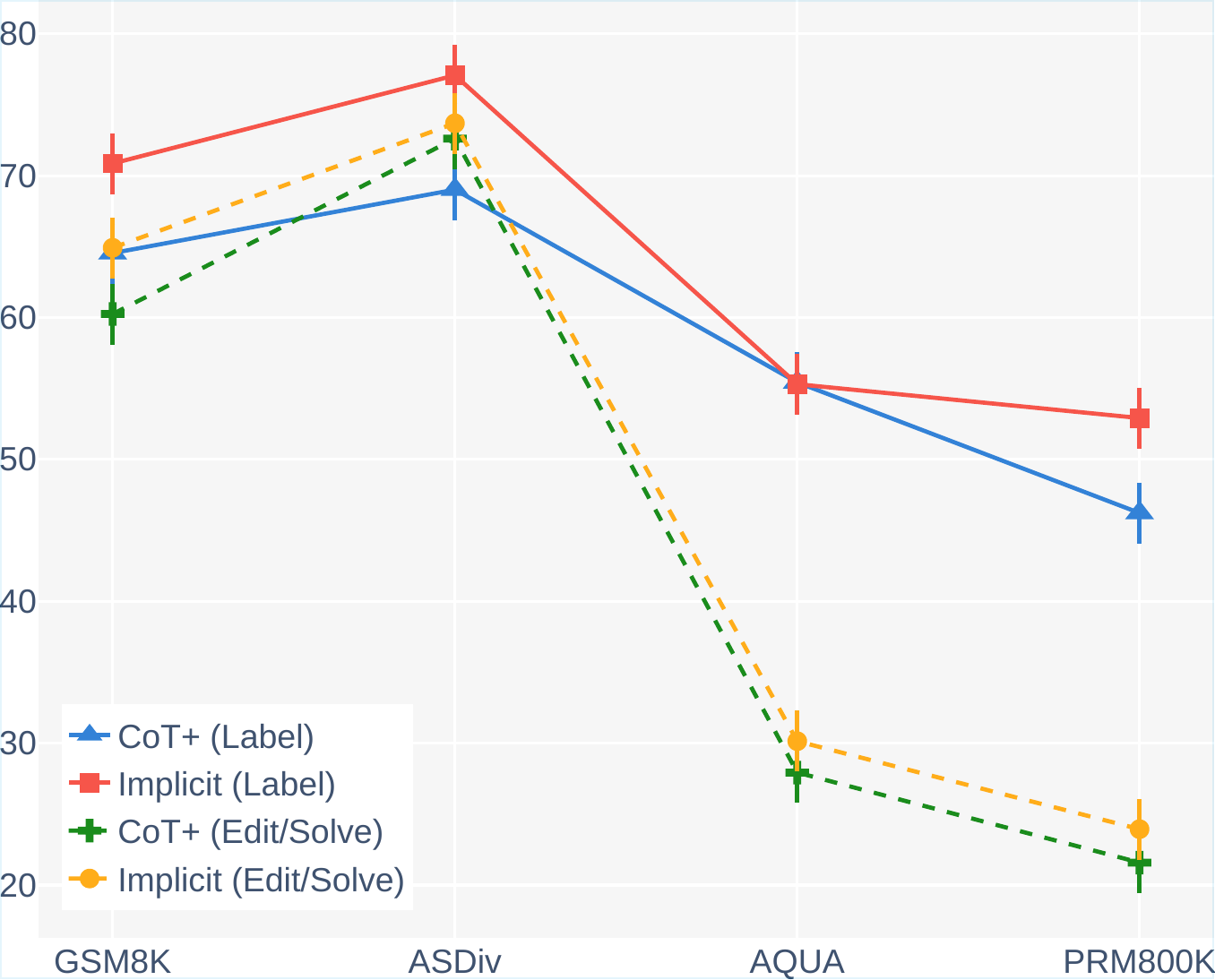}
\caption{Scores per dataset of CoT+ and implicit prompting for (a) the weighted F1-score of the labelling task, and (b) the averaged accuracy across the editing and solving tasks. Scores are averaged across all LLMs.}
\label{fig:cot_plus_results}
\vspace{-2pt}
\end{figure}

\paragraph{CoT+} extends the context by increasing the number of examples in our CoT prompt from eight to fourteen. Additional examples are randomly selected from an identical distribution to the original eight examples. We compare this setup to implicit learning prompts containing eight few-shot examples as in our standard experimental setting. The addition of six new in-context examples to the CoT prompt results in an approximately equal context length between the two settings. It also constitutes a particularly strong baseline, since the new examples may provide the model with additional, novel scenarios to learn from. Figure~\ref{fig:cot_plus_results} summarises the results of this comparison (results are shown in full in Table~\ref{tab:analysis_result}). With the exception of the AQuA labelling task, where the two methods perform equally, the average performance of implicit learning is above that of CoT+ on all datasets and tasks, often substantially (3+\% improvement). This demonstrates that, at equal context length, the addition of incorrect answers is more beneficial for LLMs than the inclusion of additional diverse and valid question-answer pairs.
\paragraph{CoT-2} provides two correct CoT answers for each in-context question. We show the differences in performance between CoT-2 and implicit learning, computed on Command R+, in Table~\ref{tab:preliminary_result}. We note that implicit learning outperforms CoT-2 in almost all cases, with average performance gains of +1.5\% in the labelling task (measured in F1-score) and +4.4\% in the editing and solving tasks (measured in accuracy). These results indicate that LLMs prompted for implicit learning gain a better understanding of the patterns that inform correct answers—and how these differ from incorrect answers—which prompting with only correct reasoning traces may not sufficiently elicit.

\subsection{Analysis of Generated Rationales}\label{sec:analysis}
\begin{table}[!t]
\centering
\renewcommand\arraystretch{1.0}
\setlength{\tabcolsep}{7.2pt}
\scalebox{0.9}{
\begin{tabular}{L{2.6cm}r}
\hline
\toprule
\textbf{Strategy} 
& \textbf{\makecell[r]{Avg. human evaluation score}} \\
\midrule
CoT
& $0.68$  \\
Explicit
& $1.01$ \\
Implicit
& $0.98$  \\
\bottomrule
\hline
\end{tabular}
}
\caption{Human evaluation of model-generated rationales. Rationales produced with explicit and implicit learning prompts obtain similar overall evaluations. CoT prompting results in substantially worse-quality rationales being generated.
% .
}
\label{tab:humaneval_result}
\vspace{6pt}
\end{table}
\begin{table}[!t]
\centering
\renewcommand\arraystretch{1.0}
\setlength{\tabcolsep}{7.2pt}
\scalebox{0.90}{
\begin{tabular}{L{3.6cm}r}
\hline
\toprule
\textbf{Strategy} & 
\textbf{Avg. $\mathbf{n}$-gram similarity} \\
\midrule
{CoT}
& $0.086$ \\

{Explicit}
& $0.152$ \\

{Implicit}
& $0.093$ \\
\bottomrule
\hline
\end{tabular}
}
\caption{
Average $n$-gram similarity between exemplar and generated rationales for each prompting strategy.
}
\label{tab:rat_sim}
\end{table}
A follow-up research question aims to investigate how incorporating error information affects model outputs. We hypothesise that if the LLMs are incorporating error signal implicitly to improve reasoning, this should also be reflected in the downstream generated rationales. We thus ask the models to generate rationales for new incorrect answers under each prompting strategy. We assess and compare them through human evaluation, and examine whether and to what extent LLMs overfit to in-context rationales when these are provided. Finally, we inspect all rationales visually and give an overview of their representative characteristics.\vspace{4pt}

\paragraph{Rationale quality.} To ascertain whether, and to what extent, LLMs infer implicit information between incorrect and correct answers with different prompting strategies, we carry out a blind human evaluation study of rationales generated using distinct prompts. We randomly select 300 rationales generated by running the answer labelling task on GSM8K. We select 100 rationales for each prompting strategy (CoT, explicit learning, implicit learning), and have four annotators with domain expertise score them as \textit{0--Poor}, \textit{1--Fair} or \textit{2--Good}. Table \ref{tab:humaneval_result} illustrates the average human evaluation scores achieved under each prompting strategy. We observe that CoT's performance is considerably lower than either explicit or implicit learning, with an average score of 0.68. The performance of explicit and implicit learning is similar (1.01 and 0.98 respectively). It is noteworthy that rationales generated with implicit learning prompts achieve an average score that is within only 0.03 of that achieved by explicit learning. This is evidence that LLMs can infer high-quality corrective rationales implicitly, simply observing correct and incorrect answers side by side, and that the effect of adding example rationales to the context is negligible. To understand how much annotators agree with each other when assessing the different prompting strategies, we measure the proportion of ‘poor’, ‘fair’, and ‘good’ qualitative labels assigned by each annotator to rationales generated with CoT, explicit and implicit prompting respectively. For each prompting method, we compute the median absolute deviation (MAD) of each label across all annotators. We find the MAD to be within $0.05$ in all cases, indicating that annotators largely agree with each other on the overall quality of the rationales generated with each prompting strategy, and the proportions of qualitative labels they assign to the rationales obtained with a given strategy are fairly similar. In Appendix~\ref{sec:human_eval}, we show the fine-grained proportions of labels assigned by the annotators to each strategy.

\paragraph{Rationale similarity.} A plausible reason why learning with explicit corrective feedback underperforms implicit learning is that LLMs may be over-constrained by the rationales. To validate this hypothesis, we investigate how similar the new rationales generated by the models are to the in-context ones. As shown in Table~\ref{tab:rat_sim}, the average $n$-gram similarity score ($n=2$) of rationales generated by LLMs prompted for explicit learning is substantially higher than that obtained with rationales output with the other methods (note that the other methods do not include exemplar rationales in the context). Rouge-1 and Rouge-L scores, shown in Appendix~\ref{sec:rouge}, follow a similar trend. It thus appears that LLMs tend to copy the patterns in the exemplar rationales when these are provided. This suggests that overfitting may be responsible for the lower performance of explicit learning.\vspace{4pt}

\promptbottom{rat:cot}{CoT}{The provided answer is correct as it solves the word problem correctly by letting the number of fish in each aquarium be x. The answer follows the narrative of the problem and uses the information that the difference in snails between the two aquariums is twice the amount of fish in both aquariums. The solution is coherent and arrives at the conclusion that each aquarium has 14 fish.\label{rat:cot}}
\promptbottom{rat:exp}{Explicit}{The provided answer is incorrect because it does not actually provide a numerical value for the number of fish in each aquarium, which is what the question is asking for. Instead, it repeats the expression "Let x be the number of fish in each aquarium" multiple times, which is not a valid answer. The answer also does not explain how the problem's conditions are reflected in the solution, which is twice the number of fish in each aquarium is equal to the difference in the number of snails between the two aquariums.\label{rat:exp}}
\promptbottom{rat:imp}{Implicit}{The provided answer is incorrect because the solution fails to provide the final calculation to determine the number of fish in each aquarium. The answer assumes the role of x as the number of fish in each aquarium but does not conclude the equation.\label{rat:imp}}

\paragraph{Rationale length and appearance.} We inspect the generated rationales and find that those generated via explicit learning prompting tend to be more verbose. As a representative example, consider the math reasoning problem \textit{“There are
4 snails in one aquarium and 32 snails in another
aquarium. The difference between the number of
snails in the two aquariums is twice the amount of
fish in both aquariums. If both aquariums have the
same number of fish in them, how many fish are
there in each aquarium?”}. We take an incorrect,
model-generated answer to this problem which assigns the unknown number of fish to the variable
$x$ but does not proceed to solve for $x$. We show
three representative rationales generated for this
question-answer pair using the three prompting
strategies: CoT (\hyperref[rat:cot]{Rationale 1}), explicit learning (\hyperref[rat:exp]{Rationale 2}) and implicit learning (\hyperref[rat:imp]{Rationale 3}). We
observe that the rationale produced using CoT fails
to identify the error. It also hallucinates that the
number of fish in each aquarium is 14, which is neither the correct solution nor a value that appears in
the answer. In contrast, prompting for both explicit and
implicit learning produces accurate rationales.
Note that the latter—generated without exemplar
rationales in the context to use as guidelines—is
more succinct yet equally exhaustive. Indeed, we observe that rationales generated via explicit learning prompting are substantially longer on average (373 average length, in characters), similar to those shown in-context (423 average length), which further confirms the overfitting hypothesis. In contrast, rationales produced with implicit learning prompting are over a third shorter (237 average length).

\section{Conclusion}

We have investigated in-context \textit{implicit} learning from mistakes across a range of LLM families and sizes, and found that it outperforms both chain-of-thought prompting and \textit{explicit} learning in challenging math reasoning tasks. Our analysis shows that although incorrect answers benefit LLMs more than additional correct ones, providing explicit corrective feedback limits those advantages, as models tend to overfit to it. Our findings are as noteworthy as they are surprising, since they call into question the benefits of widely used corrective rationales to aid LLMs in learning from mistakes. These rationales are prevalent in current frameworks despite being expensive to curate at scale, yet our investigation suggests that they are redundant and can even hurt performance by adding unnecessary constraints.

\section*{Limitations}

We have carried out an exhaustive investigation of implicit learning from mistakes, focused on in-context learning. It is worth noting that implicit learning examples---which consist of triples of the form \textit{(question, incorrect answer, correct answer)}---can be obtained at scale by simply running more and less capable LLMs on training set questions. This opens up the possibility of investigating performance differences between explicit and implicit learning also in other paradigms, such as in the fine-tuning setting. Future work can investigate whether the results established in this paper extend to models fine-tuned using similar strategies.

In our experiments, we use four datasets covering different math topics and difficulty levels, extract multiple subtasks from each dataset, use four prompting strategies (CoT, CoT+, implicit, explicit), and seven LLMs. This totals 364 distinct experimental setups, each run five times for robustness. Given the extensiveness of our experiments, it was infeasible to explore further domains other than math reasoning within the scope of this work. Math benchmarks were chosen as a reliable proxy for LLM reasoning in accordance with established prior literature \cite{ahn-etal-2024-large, paul-etal-2024-refiner, ruis2024proceduralknowledgepretrainingdrives, liu2025llmscapablestablereasoning}. While there exists prior work investigating related topics with greater breadth \cite{lampinen-etal-2022-language}, we leave similar investigations in other domains to future work.

\section*{Ethical Considerations}

This study relies solely on established, publicly available math reasoning benchmarks and focuses on evaluating different prompting strategies. As such, it does not involve sensitive data or foreseeable ethical risks. Our use of the models and datasets described in this paper complies with all applicable licenses. 

%%%%%%%%%%%%%%%%%%%%%%

\bibliography{custom}\clearpage

%%%%%%%%%%%%%%%%%%%%%%

\appendix

\section{Prompts}\label{sec:prompts}

All prompts are of the form
\[ \texttt{[examples]\textbackslash n\textbackslash n\textbackslash n\textbackslash n[instruction]}, \]
where the \textit{examples} are specific to the prompting strategy (i.e., CoT, explicit learning, implicit learning), while the \textit{instruction} is task-dependent (i.e., labelling an answer or step, editing an incorrect answer, solving a new question). In the next sections, we illustrate the examples for each prompting strategy and the instructions for each task.

\subsection{Examples}

All few-shot examples shown below are for GSM8K. Examples for all datasets are included at \url{https://github.com/lisaalaz/implicit-learning-with-llms}.

%%%%%%%%%%%%%%%%%%%%%%%%%%%%

\paragraph{CoT Examples.}\prompttext{Below you will find some questions followed by the answer.}

\vspace{12pt}

\noindent{\prompttext{Question: A class has 32 students and they all need to present their projects. Every period is 40 minutes long. How many periods will it take for every student to present their project if they have 5 minutes to do so?}}

\vspace{6pt}

\noindent{\prompttext{Answer: There are 32 students and each student needs 5 minutes to present their project.\\
So the total time needed is 32 x 5 = 160 minutes.\\
There are 40 minutes in each period.\\
So the number of periods needed is 160/40 = 4.\\
So, the final answer is 4.}}

\vspace{12pt}

\noindent{\prompttext{Question: A bag contains 50 fewer baseball cards than football cards. There are 4 times as many football cards as hockey cards. If there are 200 hockey cards in the bag, how many cards are there altogether?}}

\vspace{6pt}

\noindent{\prompttext{Answer: There are 4 times as many football cards as hockey cards, so there are 4 x 200 = 800 football cards.\\
There are 50 fewer baseball cards than football cards, so there are 800 - 50 = 750 baseball cards.\\
There are 200 hockey cards, 800 football cards and 750 baseball cards for a total of 200 + 800 + 750 = 1750 cards.\\
So, the final answer is 1750.}}

\vspace{12pt}

\noindent{\prompttext{Question: A bag of caramel cookies has 20 cookies inside and a box of cookies has 4 bags in total. How many calories are inside the box if each cookie is 20 calories?}}

\vspace{6pt}

\noindent{\prompttext{Answer: There are 20 cookies in each bag and 4 bags in total.\\
So there are 20 x 4 = 80 cookies in total.\\
Each cookie is 20 calories, so the total calories are 80 x 20 = 1600 calories.\\
So, the final answer is 1600.}}

\vspace{12pt}

\noindent{\prompttext{Question: There are four members in one household. Each member consumes 3 slices of bread during breakfast and 2 slices of bread for snacks. A loaf of bread has 12 slices. How many days will five loaves of bread last in this family?}}

\vspace{6pt}

\noindent{\prompttext{Answer: Each member consumes 3+2 = 5 slices of bread per day.\\
There are 4 members so they consume 5 x 4 = 20 slices of bread per day.\\
Each loaf has 12 slices so 5 loaves have 5 x 12 = 60 slices.\\
60 slices of bread will last 60/20 = 3 days.\\
So, the final answer is 3.}}

\vspace{12pt}

\noindent{\prompttext{Question: Bill decides to bring donuts to work for a meeting that day. He buys a box of donuts that has 50 in total in them. He eats 2 on the ride in because he's hungry. When he gets to the office, the secretary takes another 4 out of the box when he's not looking. Lastly, right before the meeting Bill sets the box down on his desk and leaves the office to take a phone call. While Bill's away, his coworkers steal half the remaining donuts. Bill comes back and brings the box into his meeting. How many donuts are left in the box?}}

\vspace{6pt}

\noindent{\prompttext{Answer: Bill eats 2 donuts on the ride in, leaving 50 - 2 = 48 donuts.\\
The secretary eats 4 donuts, leaving 48 - 4 = 44 donuts.\\
Bill's coworkers eat half of the remaining donuts, which is 44/2 = 22 donuts.\\
So Bill has 44 - 22 = 22 donuts left.\\
So, the final answer is 22.}}

\vspace{12pt}

\noindent{\prompttext{Question: Boris has 100 pieces of Halloween candy. His daughter eats 8 pieces of candy. He separates the remaining pieces of candy into equal portions into 4 different bowls. Then he takes away 3 pieces of candy from each bowl to keep for himself. How many pieces of candy are in one bowl?}}

\vspace{6pt}

\noindent{\prompttext{Answer: Boris has 100 - 8 = 92 pieces of candy left.\\
He divides the 92 pieces of candy into 4 bowls. 92/4 = 23 pieces of candy in each bowl.\\
Then he takes away 3 pieces of candy from each bowl. 23 - 3 = 20 pieces of candy in each bowl.\\
So, the final answer is 20.}}

\vspace{12pt}

\noindent{\prompttext{Question: Yvonne and Janna were writing their 1000-word pair research paper. Yvonne was able to write 400 words while Janna wrote 150 more words than Yvonne. When they edited their paper, they removed 20 words and added twice as many words as they removed. How many more words should they add to reach the research paper requirement?}}

\vspace{6pt}

\noindent{\prompttext{Answer: Yvonne wrote 400 words and Janna wrote 150 more words than Yvonne, so Janna wrote 400 + 150 = 550 words.\\
Together they wrote 400 + 550 = 950 words.\\
After editing, they removed 20 words and added 2 x 20 = 40 words. So they have 950 - 20 + 40 = 970 words.\\
They need 1000 words, so they need 1000 - 970 = 30 more words.\\
So, the final answer is 30.}}

\vspace{12pt}

\noindent{\prompttext{Question: Luke started working on a 1000-piece jigsaw puzzle. The first day he worked on it, he put together 10 percent of the pieces. On the second day, he put together another 20 percent of the remaining pieces from the first day. On the third day, he put together 30 percent of the remaining pieces from the previous day. How many pieces does he have left to complete after the third day?}}

\vspace{6pt}

\noindent{\prompttext{Answer: 10 percent of 1000 is 100.\\
So he has 100 pieces done after the first day.
He has 900 pieces left.\\
20 percent of 900 is 180.\\
So he has 100 + 180 = 280 pieces done after the second day.\\
He has 720 pieces left.\\
30 percent of 720 is 216.\\
So he has 280 + 216 = 496 pieces done after the third day.\\
He has 720 - 216 = 504 pieces left.\\
So, the final answer is 504.}}

%%%%%%%%%%%%%%%%%%%%%%

\paragraph{Explicit Learning Examples.}\prompttext{Below you will find some questions followed by a wrong answer, a rationale and the correct answer.}

\vspace{12pt}

\noindent{\prompttext{Question: A class has 32 students and they all need to present their projects. Every period is 40 minutes long. How many periods will it take for every student to present their project if they have 5 minutes to do so?}}

\vspace{6pt}

\noindent{\prompttext{Wrong answer: 32 students x 5 minutes = 160 minutes.\\
5 minutes x 40 minutes = 200 minutes.\\
160 minutes + 200 minutes = 360 minutes.\\
So it will take 360 minutes or 6 hours for every student to present their project.\\
So, the final answer is 360.}}

\vspace{6pt}

\noindent{\prompttext{Rationale: The error occurs in the second line of the wrong answer. The calculation of 5 minutes x 40 minutes = 200 minutes is incorrect and irrelevant to the problem. The correct approach is to calculate the total time needed for all presentations (32 students x 5 minutes each = 160 minutes) and then divide this by the length of each period (40 minutes) to find the number of periods needed. The incorrect answer also incorrectly converts the total minutes into hours, which is not asked for in the question.}}

\vspace{6pt}

\noindent{\prompttext{Correct answer: There are 32 students and each student needs 5 minutes to present their project.\\
So the total time needed is 32 x 5 = 160 minutes.\\
There are 40 minutes in each period.\\
So the number of periods needed is 160/40 = 4.\\
So, the final answer is 4.}}

\vspace{12pt}

\noindent{\prompttext{Question: A bag contains 50 fewer baseball cards than football cards. There are 4 times as many football cards as hockey cards. If there are 200 hockey cards in the bag, how many cards are there altogether?}}

\vspace{6pt}

\noindent{\prompttext{Wrong answer: 50 fewer baseball cards than football cards.\\
There are 4 times as many football cards as hockey cards.\\
If there are 200 hockey cards in the bag, there are 200 x 4 = 800 football cards and 800 - 50 = 750 baseball cards.\\
So there are 750 + 800 = 1550 cards altogether.\\
So, the final answer is 1550.}}

\vspace{6pt}

\noindent{\prompttext{Rationale: The error occurs in the last line of the wrong answer. The wrong answer incorrectly calculates the total number of cards. The total number of cards should be the sum of the hockey cards, football cards, and baseball cards. The correct calculation is 200 (hockey cards) + 800 (football cards) + 750 (baseball cards) = 1750 cards.}}

\vspace{6pt}

\noindent{\prompttext{Correct answer: There are 4 times as many football cards as hockey cards, so there are 4 x 200 = 800 football cards.\\
There are 50 fewer baseball cards than football cards, so there are 800 - 50 = 750 baseball cards.\\
There are 200 hockey cards, 800 football cards and 750 baseball cards for a total of 200 + 800 + 750 = 1750 cards.\\
So, the final answer is 1750.}}

\vspace{12pt}

\noindent{\prompttext{Question: A bag of caramel cookies has 20 cookies inside and a box of cookies has 4 bags in total. How many calories are inside the box if each cookie is 20 calories?}}

\vspace{6pt}

\noindent{\prompttext{Wrong answer: 20 cookies in a bag amount to 20 x 20 = 400 calories.\\
There are 4 bags in a box.\\
So 400/4 = 100 calories in a box.\\
So, the final answer is 100.}}

\vspace{6pt}

\noindent{\prompttext{Rationale: The error occurs in line 3 of the wrong answer. The incorrect part of this line is the calculation of the total calories in a box. The total calories should be calculated by multiplying the total number of cookies in the box (which is 80, not 20) by the calories per cookie. The wrong answer incorrectly divided the total calories in a bag by the number of bags in a box.}}

\vspace{6pt}

\noindent{\prompttext{Correct answer: There are 20 cookies in each bag and 4 bags in total.\\
So there are 20 x 4 = 80 cookies in total.\\
Each cookie is 20 calories, so the total calories are 80 x 20 = 1600 calories.\\
So, the final answer is 1600.}}

\vspace{12pt}

\noindent{\prompttext{Question: There are four members in one household. Each member consumes 3 slices of bread during breakfast and 2 slices of bread for snacks. A loaf of bread has 12 slices. How many days will five loaves of bread last in this family?}}

\vspace{6pt}

\noindent{\prompttext{Wrong answer: 12/4 = 3 slices of bread per person per day.\\
5 loaves of bread = 5 x 12 = 60 slices of bread.\\
60/4 = 15 days.\\
So, the final answer is 15.}}

\vspace{6pt}

\noindent{\prompttext{Rationale: The error occurs in the first line of the wrong answer. The incorrect part of this line is the calculation of the slices of bread consumed per person per day. The problem states that each member consumes 3 slices of bread for breakfast and 2 slices for snacks, so each member consumes a total of 5 slices per day, not 3. Therefore, the total slices of bread consumed per day by the family should be 5 slices per person * 4 people = 20 slices, not 12. The correct calculation should then be 60 slices / 20 slices per day = 3 days.}}

\vspace{6pt}

\noindent{\prompttext{Correct answer: Each member consumes 3+2 = 5 slices of bread per day.\\
There are 4 members so they consume 5 x 4 = 20 slices of bread per day.\\
Each loaf has 12 slices so 5 loaves have 5 x 12 = 60 slices.\\
60 slices of bread will last 60/20 = 3 days.\\
So, the final answer is 3.}}

\vspace{12pt}

\noindent{\prompttext{Question: Bill decides to bring donuts to work for a meeting that day. He buys a box of donuts that has 50 in total in them. He eats 2 on the ride in because he's hungry. When he gets to the office, the secretary takes another 4 out of the box when he's not looking. Lastly, right before the meeting Bill sets the box down on his desk and leaves the office to take a phone call. While Bill's away, his coworkers steal half the remaining donuts. Bill comes back and brings the box into his meeting. How many donuts are left in the box?}}

\vspace{6pt}

\noindent{\prompttext{Wrong answer: Bill eats 2 donuts on the way to work.\\
The secretary takes 4 donuts.\\
Bill's coworkers steal 25 donuts.\\
Bill brings the box into the meeting with 23 donuts.\\
So, the final answer is 23.}}

\vspace{6pt}

\noindent{\prompttext{Rationale: The error occurs in line 2 of the wrong answer. The incorrect part of this line is the calculation of the number of donuts stolen by Bill's coworkers. The problem states that his coworkers steal half of the remaining donuts, not half of the original number of donuts. Therefore, the number of donuts stolen by his coworkers should be half of the remaining 44 donuts, not half of the original 50 donuts.}}

\vspace{6pt}

\noindent{\prompttext{Correct answer: Bill eats 2 donuts on the ride in, leaving 50 - 2 = 48 donuts.\\
The secretary eats 4 donuts, leaving 48 - 4 = 44 donuts.\\
Bill's coworkers eat half of the remaining donuts, which is 44/2 = 22 donuts.\\
So Bill has 44 - 22 = 22 donuts left.\\
So, the final answer is 22.}}

\vspace{12pt}

\noindent{\prompttext{Question: Boris has 100 pieces of Halloween candy. His daughter eats 8 pieces of candy. He separates the remaining pieces of candy into equal portions into 4 different bowls. Then he takes away 3 pieces of candy from each bowl to keep for himself. How many pieces of candy are in one bowl?}}

\vspace{6pt}

\noindent{\prompttext{Wrong answer: Boris has 100 pieces of candy.\\
His daughter eats 8 pieces of candy.\\
He separates the remaining pieces of candy into equal portions into 4 different bowls.\\
Then he takes away 3 pieces of candy from each bowl to keep for himself. So he has 100-8 = 92 pieces of candy in 4 bowls.\\
He takes away 3 pieces of candy from each bowl. So he has 92-3 = 89 pieces of candy in 4 bowls.\\
So, the final answer is 89.}}

\vspace{6pt}

\noindent{\prompttext{Rationale: The error occurs at line 5 of the wrong answer. The incorrect part of this line is the calculation of the number of candies left after Boris takes away 3 pieces from each bowl. The subtraction should be done after dividing the total number of candies by 4 (the number of bowls), not before. Also, the subtraction should be 3*4=12, not just 3, because Boris takes away 3 pieces from each of the 4 bowls.}}

\vspace{6pt}

\noindent{\prompttext{Correct answer: Boris has 100 - 8 = 92 pieces of candy left.\\
He divides the 92 pieces of candy into 4 bowls. 92/4 = 23 pieces of candy in each bowl.\\
Then he takes away 3 pieces of candy from each bowl. 23 - 3 = 20 pieces of candy in each bowl.\\
So, the final answer is 20.}}

\vspace{12pt}

\noindent{\prompttext{Question: Yvonne and Janna were writing their 1000-word pair research paper. Yvonne was able to write 400 words while Janna wrote 150 more words than Yvonne. When they edited their paper, they removed 20 words and added twice as many words as they removed. How many more words should they add to reach the research paper requirement?}}

\vspace{6pt}

\noindent{\prompttext{Wrong answer: Yvonne wrote 400 words and Janna wrote 150 more words than Yvonne. So Janna wrote 550 words.\\
They removed 20 words and added twice as many words as they removed. So they added 40 words.\\
They need to add 40 words to reach the research paper requirement.\\
So, the final answer is 40.}}

\vspace{6pt}

\noindent{\prompttext{Rationale: The error occurs in line 3 of the wrong answer. The incorrect part of this line is the calculation of the total words they have after editing. They should subtract the words they removed and add the words they added to the total words they initially wrote. The wrong answer also incorrectly calculated the number of words they need to add to reach the research paper requirement. They should subtract the total words they have after editing from the research paper requirement.}}

\vspace{6pt}

\noindent{\prompttext{Correct answer: Yvonne wrote 400 words and Janna wrote 150 more words than Yvonne, so Janna wrote 400 + 150 = 550 words.\\
Together they wrote 400 + 550 = 950 words.\\
After editing, they removed 20 words and added 2 x 20 = 40 words. So they have 950 - 20 + 40 = 970 words.\\
They need 1000 words, so they need 1000 - 970 = 30 more words.\\
So, the final answer is 30.}}

\vspace{12pt}

\noindent{\prompttext{Question: Luke started working on a 1000-piece jigsaw puzzle. The first day he worked on it, he put together 10 percent of the pieces. On the second day, he put together another 20 percent of the remaining pieces from the first day. On the third day, he put together 30 percent of the remaining pieces from the previous day. How many pieces does he have left to complete after the third day?}}

\vspace{6pt}

\noindent{\prompttext{Wrong answer: 1000 pieces = 1000/100 = 10 pieces Luke put together 10 pieces on the first day.\\
He put together 20 pieces on the second day.\\
He put together 30 pieces on the third day.\\
So he has 10+20+30 = 60 pieces left to complete after the third day.\\
So, the final answer is 60.}}

\vspace{6pt}

\noindent{\prompttext{Rationale: The error occurs in the first line of the wrong answer. The wrong answer incorrectly calculates 10 percent of 1000 as 10 pieces, when it should be 100 pieces. The same mistake is made for the calculations on the second and third day. The correct way to solve this problem is to calculate the percentage of the remaining pieces each day, not a percentage of the original 1000 pieces.}}

\vspace{6pt}

\noindent{\prompttext{Correct answer: 10 percent of 1000 is 100.\\
So he has 100 pieces done after the first day.
He has 900 pieces left.\\
20 percent of 900 is 180.\\
So he has 100 + 180 = 280 pieces done after the second day.\\
He has 720 pieces left.\\
30 percent of 720 is 216.\\
So he has 280 + 216 = 496 pieces done after the third day.\\
He has 720 - 216 = 504 pieces left.\\
So, the final answer is 504.}}

%%%%%%%%%%%%%%%%%%%%%%%%%%%%

\paragraph{Implicit Learning Examples.}\prompttext{Below you will find some questions followed by a wrong answer and the correct answer.}

\vspace{12pt}

\noindent{\prompttext{Question: A class has 32 students and they all need to present their projects. Every period is 40 minutes long. How many periods will it take for every student to present their project if they have 5 minutes to do so?}}

\vspace{6pt}

\noindent{\prompttext{Wrong answer: 32 students x 5 minutes = 160 minutes.\\
5 minutes x 40 minutes = 200 minutes.\\
160 minutes + 200 minutes = 360 minutes.\\
So it will take 360 minutes or 6 hours for every student to present their project.\\
So, the final answer is 360.}}

\vspace{6pt}

\noindent{\prompttext{Correct answer: There are 32 students and each student needs 5 minutes to present their project.\\
So the total time needed is 32 x 5 = 160 minutes.\\
There are 40 minutes in each period.\\
So the number of periods needed is 160/40 = 4.\\
So, the final answer is 4.}}

\vspace{12pt}

\noindent{\prompttext{Question: A bag contains 50 fewer baseball cards than football cards. There are 4 times as many football cards as hockey cards. If there are 200 hockey cards in the bag, how many cards are there altogether?}}

\vspace{6pt}

\noindent{\prompttext{Wrong answer: 50 fewer baseball cards than football cards.\\
There are 4 times as many football cards as hockey cards.\\
If there are 200 hockey cards in the bag, there are 200 x 4 = 800 football cards and 800 - 50 = 750 baseball cards.\\
So there are 750 + 800 = 1550 cards altogether.\\
So, the final answer is 1550.}}

\vspace{6pt}

\noindent{\prompttext{Correct answer: There are 4 times as many football cards as hockey cards, so there are 4 x 200 = 800 football cards.\\
There are 50 fewer baseball cards than football cards, so there are 800 - 50 = 750 baseball cards.\\
There are 200 hockey cards, 800 football cards and 750 baseball cards for a total of 200 + 800 + 750 = 1750 cards.\\
So, the final answer is 1750.}}

\vspace{12pt}

\noindent{\prompttext{Question: A bag of caramel cookies has 20 cookies inside and a box of cookies has 4 bags in total. How many calories are inside the box if each cookie is 20 calories?}}

\vspace{6pt}

\noindent{\prompttext{Wrong answer: 20 cookies in a bag amount to 20 x 20 = 400 calories.\\
There are 4 bags in a box.\\
So 400/4 = 100 calories in a box.\\
So, the final answer is 100.}}

\vspace{6pt}

\noindent{\prompttext{Correct answer: There are 20 cookies in each bag and 4 bags in total.\\
So there are 20 x 4 = 80 cookies in total.\\
Each cookie is 20 calories, so the total calories are 80 x 20 = 1600 calories.\\
So, the final answer is 1600.}}

\vspace{12pt}

\noindent{\prompttext{Question: There are four members in one household. Each member consumes 3 slices of bread during breakfast and 2 slices of bread for snacks. A loaf of bread has 12 slices. How many days will five loaves of bread last in this family?}}

\vspace{6pt}

\noindent{\prompttext{Wrong answer: 12/4 = 3 slices of bread per person per day.\\
5 loaves of bread = 5 x 12 = 60 slices of bread.\\
60/4 = 15 days.\\
So, the final answer is 15.}}

\vspace{6pt}

\noindent{\prompttext{Correct answer: Each member consumes 3+2 = 5 slices of bread per day.\\
There are 4 members so they consume 5 x 4 = 20 slices of bread per day.\\
Each loaf has 12 slices so 5 loaves have 5 x 12 = 60 slices.\\
60 slices of bread will last 60/20 = 3 days.\\
So, the final answer is 3.}}

\vspace{12pt}

\noindent{\prompttext{Question: Bill decides to bring donuts to work for a meeting that day. He buys a box of donuts that has 50 in total in them. He eats 2 on the ride in because he's hungry. When he gets to the office, the secretary takes another 4 out of the box when he's not looking. Lastly, right before the meeting Bill sets the box down on his desk and leaves the office to take a phone call. While Bill's away, his coworkers steal half the remaining donuts. Bill comes back and brings the box into his meeting. How many donuts are left in the box?}}

\vspace{6pt}

\noindent{\prompttext{Wrong answer: Bill eats 2 donuts on the way to work.\\
The secretary takes 4 donuts.\\
Bill's coworkers steal 25 donuts.\\
Bill brings the box into the meeting with 23 donuts.\\
So, the final answer is 23.}}

\vspace{6pt}

\noindent{\prompttext{Correct answer: Bill eats 2 donuts on the ride in, leaving 50 - 2 = 48 donuts.\\
The secretary eats 4 donuts, leaving 48 - 4 = 44 donuts.\\
Bill's coworkers eat half of the remaining donuts, which is 44/2 = 22 donuts.\\
So Bill has 44 - 22 = 22 donuts left.\\
So, the final answer is 22.}}

\vspace{12pt}

\noindent{\prompttext{Question: Boris has 100 pieces of Halloween candy. His daughter eats 8 pieces of candy. He separates the remaining pieces of candy into equal portions into 4 different bowls. Then he takes away 3 pieces of candy from each bowl to keep for himself. How many pieces of candy are in one bowl?}}

\vspace{6pt}

\noindent{\prompttext{Wrong answer: Boris has 100 pieces of candy.\\
His daughter eats 8 pieces of candy.\\
He separates the remaining pieces of candy into equal portions into 4 different bowls.\\
Then he takes away 3 pieces of candy from each bowl to keep for himself. So he has 100-8 = 92 pieces of candy in 4 bowls.\\
He takes away 3 pieces of candy from each bowl. So he has 92-3 = 89 pieces of candy in 4 bowls.\\
So, the final answer is 89.}}

\vspace{6pt}

\noindent{\prompttext{Correct answer: Boris has 100 - 8 = 92 pieces of candy left.\\
He divides the 92 pieces of candy into 4 bowls. 92/4 = 23 pieces of candy in each bowl.\\
Then he takes away 3 pieces of candy from each bowl. 23 - 3 = 20 pieces of candy in each bowl.\\
So, the final answer is 20.}}

\vspace{12pt}

\noindent{\prompttext{Question: Yvonne and Janna were writing their 1000-word pair research paper. Yvonne was able to write 400 words while Janna wrote 150 more words than Yvonne. When they edited their paper, they removed 20 words and added twice as many words as they removed. How many more words should they add to reach the research paper requirement?}}

\vspace{6pt}

\noindent{\prompttext{Wrong answer: Yvonne wrote 400 words and Janna wrote 150 more words than Yvonne. So Janna wrote 550 words.\\
They removed 20 words and added twice as many words as they removed. So they added 40 words.\\
They need to add 40 words to reach the research paper requirement.\\
So, the final answer is 40.}}

\vspace{6pt}

\noindent{\prompttext{Correct answer: Yvonne wrote 400 words and Janna wrote 150 more words than Yvonne, so Janna wrote 400 + 150 = 550 words.\\
Together they wrote 400 + 550 = 950 words.\\
After editing, they removed 20 words and added 2 x 20 = 40 words. So they have 950 - 20 + 40 = 970 words.\\
They need 1000 words, so they need 1000 - 970 = 30 more words.\\
So, the final answer is 30.}}

\vspace{12pt}

\noindent{\prompttext{Question: Luke started working on a 1000-piece jigsaw puzzle. The first day he worked on it, he put together 10 percent of the pieces. On the second day, he put together another 20 percent of the remaining pieces from the first day. On the third day, he put together 30 percent of the remaining pieces from the previous day. How many pieces does he have left to complete after the third day?}}

\vspace{6pt}

\noindent{\prompttext{Wrong answer: 1000 pieces = 1000/100 = 10 pieces Luke put together 10 pieces on the first day.\\
He put together 20 pieces on the second day.\\
He put together 30 pieces on the third day.\\
So he has 10+20+30 = 60 pieces left to complete after the third day.\\
So, the final answer is 60.}}

\vspace{6pt}

\noindent{\prompttext{Correct answer: 10 percent of 1000 is 100.\\
So he has 100 pieces done after the first day.
He has 900 pieces left.\\
20 percent of 900 is 180.\\
So he has 100 + 180 = 280 pieces done after the second day.\\
He has 720 pieces left.\\
30 percent of 720 is 216.\\
So he has 280 + 216 = 496 pieces done after the third day.\\
He has 720 - 216 = 504 pieces left.\\
So, the final answer is 504.}}

%%%%%%%%%%%%%%%%%%%%%%%%%%%

\prompt{ins:label_ans}{Answer labelling instruction}{Now apply what you have learned and determine if the answer below is wrong or correct. Begin your evaluation by providing a short explanation. Be as objective as possible. After providing your explanation, you must rate the answer as either wrong or correct by strictly following this format: "[[rating]]", for example: "Rating: [[wrong]]" or "Rating: [[correct]]". You can only use the words wrong or correct as the final rating.}

\prompt{ins:label_step}{Step labelling instruction}{Now apply what you have learned when reading the question and the step-by-step answer below. The answer may not yet be complete. Your task is to determine if the current step will lead to a wrong or correct final answer, based on the question and the previous steps. Begin your evaluation by providing a short explanation. Be as objective as possible. After providing your explanation, you must rate the current reasoning step as either wrong or correct by strictly following this format: "[[rating]]", for example: "Rating: [[wrong]]" or "Rating: [[correct]]". You can only use the words wrong or correct as the final rating.}

\prompt{ins:edit}{Editing instruction}{Now apply what you have learned and given the question below and a wrong answer, write the correct answer.}

\prompt{ins:solve}{Solving instruction}{Now apply what you have learned and answer the question below.}

\subsection{Instructions}

The instructions for the answer labelling, step labelling, editing and solving tasks are shown in I\ref{ins:label_ans}, I\ref{ins:label_step}, I\ref{ins:edit} and I\ref{ins:solve}, respectively.

\section{Models}\label{sec:models}

We list below each of the seven LLMs tested, with the corresponding API provider and model identifier. With all LLMs we use hyperparameters $t = 0$, $p = 1$, and $k = 1$.
\vspace{4pt}
\bgroup
\flushleft{
\begin{itemize}
    \item Llama 3.1 70B, \textit{Amazon Bedrock}, \texttt{meta.llama3-70b-instruct-v1:0}
    \item Titan Text G1 Express, \textit{Amazon Bedrock}, \texttt{amazon.titan-text-express-v1}
    \item Command R, \textit{Cohere}, \texttt{command-r-03-2024}
    \item Command R Refresh, \textit{Cohere}, \texttt{command-r-08-2024}
    \item Command R+, \textit{Cohere}, \texttt{command-r-plus-04-2024}
    \item Command R+ Refresh, \textit{Cohere}, \texttt{command-r-plus-08-2024}
    \item WizardLM, \textit{TogetherAI}, \texttt{microsoft/WizardLM-2-8x22B}
\end{itemize}
}
\egroup

\vspace{6pt}

\begin{table}[!t]
\centering
\renewcommand\arraystretch{1.2}
\setlength{\tabcolsep}{1pt}
\scalebox{0.8}{
\begin{tabular}{L{2.15cm}L{3.5cm}L{3.1cm}}

\hline
\toprule
\textbf{Dataset} 
& \textbf{\makecell[l]{Model for train\\ samples (incorrect\\answers)}} 
& \textbf{\makecell[l]{Model for test\\samples (incorrect\\+ correct answers)}}\\
\midrule
GSM8K & LLaMA 30B & Llama 2 7B \\
ASDiv
& Llama 2 7B & Llama 2 7B \\
AQuA
& Llama 3 8B & Llama 3 8B \\

\bottomrule
\hline
\end{tabular}
}
\caption{
LLMs used for answer generation.}
\label{tab:models}
\end{table}

\section{Data Preparation}\label{sec:dataprep}

\begin{table*}[t!]
\centering
\renewcommand\arraystretch{1.20}
\setlength{\tabcolsep}{4pt}
\scalebox{0.7}{
\begin{tabular}{llccccccc} % outer: 9 cols total
\toprule

% -------- TOP PART (wrapped in one cell spanning all 9) --------
\multicolumn{9}{c}{
\begin{tabular}{ll@{\extracolsep{14pt}}cccccc} % inner: 8 cols (2 label + 6 data)
{\multirow{2.2}{*}{\textbf{Model}}}
& {\multirow{2.2}{*}{\textbf{Strategy}}} 
& \multicolumn{3}{c}{\textbf{GSM8K}} 
& \multicolumn{3}{c}{\textbf{ASDiv}} \\
\cmidrule(lr){3-5} \cmidrule(lr){6-8}
& & \textbf{\texorpdfstring{label\textsubscript{ans}}} & \textbf{edit} & \textbf{solve} 
& \textbf{\texorpdfstring{label\textsubscript{ans}}} & \textbf{edit} & \textbf{solve} \\
\midrule
& CoT
& $(83.76,83.84)$ & $(81.30,81.30)$ & $(91.76,91.84)$
& $(90.08,90.12)$ & $(82.63,82.77)$ & $(90.78,90.82)$ \\
\multirow{-1.4}{*}{\textit{\makecell[l]{Llama 3 70B \\ Instruct}}}
& Explicit
& $(82.45,82.55)$ & $(84.18,84.22)$ & $(92.78,92.83)$
& $(89.98,90.02)$ & $(81.33, 81.47)$ & $\textbf{(91.49, 91.51)}$ \\
& Implicit
& $\textbf{(83.94, 84.06)}$ & $\textbf{(84.79, 84.81)}$ & $\textbf{(93.27, 93.33)}$
& $\textbf{(91.36, 91.44)}$ & $\textbf{(84.84, 84.96)}$ & $(91.08,91.12)$ \\
\cmidrule(lr){1-8}
& CoT
& $(50.43, 50.57)$ & $(17.13, 17.27)$ & $\textbf{(63.06, 63.14)}$
& $(53.31, 53.49)$ & $(49.2, 49.4)$ & $\textbf{(77.54, 77.66)}$ \\
\multirow{-1}{*}{\textit{\makecell[l]{Command R}}}
& Explicit
& $(56.93, 57.07)$ & $(24.99, 25.21)$ & $(56.62, 56.78)$
& $\textbf{(64.0, 64.2)}$ & $(47.89, 48.11)$ & $(69.52, 69.68)$ \\
& Implicit
& $\textbf{(64.13, 64.27)}$ & $\textbf{(31.00, 31.20)}$ & $(60.4, 60.6)$
& $(60.24, 60.36)$ & $\textbf{(51.34, 51.46)}$ & $(70.05, 70.15)$ \\
\cmidrule(lr){1-8}
& CoT
& $(65.74, 65.86)$ & $(47.98, 48.02)$ & $(69.66, 69.74)$
& $(78.85, 78.95)$ & $(61.81, 61.99)$ & $(81.65, 81.75)$ \\
\multirow{-1}{*}{\textit{\makecell[l]{Command R+}}}
& Explicit
& $(64.28, 64.32)$ & $(59.75, 59.85)$ & $(75.93, 76.07)$
& $(80.38, 80.42)$ & $(69.20, 69.40)$ & $(83.87, 83.93)$ \\
& Implicit
& $\textbf{(71.87, 71.93)}$ & $\textbf{(61.93, 62.07)}$ & $\textbf{(79.83, 79.97)}$
& $\textbf{(82.58, 82.62)}$ & $\textbf{(70.59, 70.81)}$ & $\textbf{(85.28, 85.32)}$ \\
\cmidrule(lr){1-8}
& CoT
& $(55.43, 55.57)$ & $(52.05, 52.15)$ & $(78.87, 78.93)$
& $(54.78, 54.82)$ & $(64.78, 64.82)$ & $(84.45, 84.55)$ \\
\multirow{-1.4}{*}{\textit{\makecell[l]{Command R \\ Refresh}}}
& Explicit
& $(48.61, 48.79)$ & $(55.87, 55.93)$ & $(75.83, 75.97)$
& $(37.85, 37.95)$ & $(69.11, 69.29)$ & $(80.85, 80.95)$ \\
& Implicit
& $\textbf{(62.42, 62.58)}$ & $\textbf{(57.34, 57.46)}$ & $\textbf{(79.14, 79.26)}$
& $\textbf{(70.31, 70.49)}$ & $\textbf{(72.18, 72.22)}$ & $\textbf{(84.76, 84.84)}$ \\
\cmidrule(lr){1-8}
& CoT
& $(46.83, 46.97)$ & $(45.84, 45.96)$ & $(75.53, 75.67)$
& $(77.69, 77.71)$ & $(78.76, 78.84)$ & $(89.37, 89.43)$ \\
\multirow{-1.4}{*}{\textit{\makecell[l]{Command R+ \\ Refresh}}}
& Explicit
& $(40.23, 40.37)$ & $(57.49, 57.71)$ & $(81.94, 82.06)$
& $(64.78, 64.82)$ & $(76.03, 76.17)$ & $(89.88, 89.92)$ \\
& Implicit
& $\textbf{(47.13, 47.27)}$ & $\textbf{(62.73, 62.87)}$ & $\textbf{(86.23, 86.37)}$
& $\textbf{(79.58, 79.62)}$ & $\textbf{(81.82, 81.98)}$ & $\textbf{(90.37, 90.43)}$ \\
\cmidrule(lr){1-8}
& CoT
& $(53.38, 53.42)$ & $(2.08, 2.12)$ & $(29.64, 29.76)$
& $(45.35, 45.45)$ & $\textbf{(13.08, 13.12)}$ & $(61.17, 61.23)$ \\
\multirow{-1.4}{*}{\textit{\makecell[l]{Titan Text G1 \\ Express}}}
& Explicit
& $(68.68, 68.72)$ & $(2.77, 2.83)$ & $(33.16, 33.24)$
& $(52.93, 53.07)$ & $(12.88, 12.92)$ & $(61.05, 61.15)$ \\
& Implicit
& $\textbf{(69.26, 69.34)}$ & $\textbf{(2.97, 3.03)}$ & $\textbf{(34.63, 34.77)}$
& $\textbf{(60.17, 60.23)}$ & $(12.89, 12.91)$ & $\textbf{(62.77, 62.83)}$ \\
\cmidrule(lr){1-8}
& CoT
& $(85.87, 85.93)$ & $(72.31, 72.49)$ & $(91.25, 91.35)$
& $(93.78, 93.82)$ & $(80.58, 80.62)$ & $(90.87, 90.93)$ \\
\multirow{-1}{*}{\textit{\makecell[l]{WizardLM}}}
& Explicit
& $(85.36, 85.44)$ & $(78.39, 78.61)$ & $\textbf{(91.43, 91.57)}$
& $(93.48, 93.52)$ & $(82.38, 82.42)$ & $\textbf{(91.39, 91.41)}$ \\
& Implicit
& $\textbf{(86.86, 86.94)}$ & $\textbf{(82.23, 82.37)}$ & $\textbf{(91.44, 91.56)}$
& $\textbf{(94.98, 95.02)}$ & $\textbf{(82.58, 82.62)}$ & $(90.58, 90.62)$ \\
\end{tabular}
} \\ % <-- end of top block, one big row
\toprule
% -------- BOTTOM PART (normal 9 columns) --------
{\multirow{2.2}{*}{\textbf{Model}}}
& {\multirow{2.2}{*}{\textbf{Strategy}}} 
& \multicolumn{3}{c}{\textbf{AQuA}}
& \multicolumn{4}{c}{\textbf{PRM800K}} \\
\cmidrule(lr){3-5} \cmidrule(lr){6-9}
& & \textbf{\texorpdfstring{label\textsubscript{ans}}} & \textbf{edit} & \textbf{solve} 
& \textbf{\texorpdfstring{label\textsubscript{ans}}} & \textbf{\texorpdfstring{label\textsubscript{step}}} & \textbf{edit} & \textbf{solve} \\
\midrule
& CoT
& $\textbf{(66.51, 66.69)}$ & $(37.27,37.33)$ & $(55.75,55.85)$
& $\textbf{(31.65, 31.75)}$ & $(49.58,49.62)$ & $(20.49,20.71)$ & $(43.83,43.97)$ \\
\multirow{-1.4}{*}{\textit{\makecell[l]{Llama 3 70B \\ Instruct}}}
& Explicit
& $(55.61,55.79)$ & $(33.91,34.09)$ & $(55.02,55.18)$
& $(18.97,19.03)$ & $(48.17,48.23)$ & $(21.67,21.93)$ & $(48.07,48.13)$ \\
& Implicit
& $(56.58,56.62)$ & $\textbf{(37.49, 37.71)}$ & $\textbf{(56.37, 56.43)}$
& $(19.18,19.22)$ & $\textbf{(49.95, 50.05)}$ & $\textbf{(26.34, 26.66)}$ & $\textbf{(48.35, 48.45)}$ \\
\cmidrule(lr){1-9}
& CoT
& $(37.03, 37.17)$ & $(7.79, 8.01)$ & $\textbf{(21.81, 21.99)}$
& $(21.34,21.46)$ & $(36.27,36.33)$ & $(4.60,4.80)$ & $(13.22,13.38)$ \\
\multirow{-1}{*}{\textit{\makecell[l]{Command R}}}
& Explicit
& $(34.07,34.33)$ & $(6.61,6.79)$ & $(17.64,17.96)$
& $(32.65,32.75)$ & $(38.98,39.02)$ & $(7.43,7.57)$ & $(12.94,13.06)$ \\
& Implicit
& $\textbf{(39.69, 39.91)}$ & $\textbf{(11.10, 11.30)}$ & $(19.08,19.12)$
& $\textbf{(55.93, 56.07)}$ & $\textbf{(43.30, 43.50)}$ & $\textbf{(8.77, 8.83)}$ & $\textbf{(14.74, 14.86)}$ \\
\cmidrule(lr){1-9}
& CoT
& $(43.73,43.87)$ & $(11.8,12.0)$ & $(31.87,32.13)$
& $(16.01,16.19)$ & $(35.77,35.83)$ & $(14.41,14.59)$ & $\textbf{(23.79, 24.01)}$ \\
\multirow{-1}{*}{\textit{\makecell[l]{Command R+}}}
& Explicit
& $(46.47,46.53)$ & $(12.38,12.62)$ & $(30.94,31.26)$
& $\textbf{(59.63, 59.77)}$ & $(38.78,38.82)$ & $(12.82,12.98)$ & $(18.05,18.15)$ \\
& Implicit
& $\textbf{(47.53, 47.67)}$ & $\textbf{(16.71, 16.89)}$ & $\textbf{(35.71, 35.89)}$
& $(59.40,59.60)$ & $\textbf{(39.18, 39.22)}$ & $\textbf{(16.53, 16.67)}$ & $(21.05,21.15)$ \\
\cmidrule(lr){1-9}
& CoT
& $(47.38,47.62)$ & $(8.42,8.58)$ & $(35.20,35.40)$
& $(68.04,68.16)$ & $(39.25,39.35)$ & $(11.64,11.76)$ & $(30.54,30.66)$ \\
\multirow{-1.4}{*}{\textit{\makecell[l]{Command R \\ Refresh}}}
& Explicit
& $(42.37,42.43)$ & $(16.46,16.54)$ & $(38.95,39.25)$
& $(67.23,67.37)$ & $\textbf{(55.85, 55.95)}$ & $\textbf{(13.03, 13.17)}$ & $(30.76,30.84)$ \\
& Implicit
& $\textbf{(50.63, 50.77)}$ & $\textbf{(16.55, 16.65)}$ & $\textbf{(40.43, 40.57)}$
& $\textbf{(71.03, 71.17)}$ & $(53.63,53.77)$ & $(11.71,11.89)$ & $\textbf{(32.04, 32.16)}$ \\
\cmidrule(lr){1-9}
& CoT
& $(60.97,61.03)$ & $\textbf{(23.44, 23.56)}$ & $(44.43, 44.57)$
& $(54.40, 54.60)$ & $\textbf{(51.85, 51.95)}$ & $(15.99, 16.21)$ & $\textbf{(31.83, 31.97)}$ \\
\multirow{-1.4}{*}{\textit{\makecell[l]{Command R+ \\ Refresh}}}
& Explicit
& $(53.56, 53.64)$ & $(53.56, 53.64)$ & $(43.06, 43.34)$
& $(73.27, 73.33)$ & $(51.63, 51.77)$ & $(15.79, 16.01)$ & $(26.77, 26.83)$ \\
& Implicit
& $\textbf{(63.20, 63.40)}$ & $(21.41, 21.59)$ & $\textbf{(47.70, 47.90)}$
& $\textbf{(74.69, 74.77)}$ & $(51.63, 51.77)$ & $\textbf{(17.41, 17.57)}$ & $(29.35, 29.45)$ \\
\cmidrule(lr){1-9}
& CoT
& $(28.14, 28.26)$ & $(0.76,0.86)$ & $(6.83,6.97)$
& $(26.53,26.67)$ & $(4.88,5.02)$ & $(1.94,2.06)$ & $(4.06,4.14)$ \\
\multirow{-1.4}{*}{\textit{\makecell[l]{Titan Text G1 \\ Express}}}
& Explicit
& $(28.89,29.01)$ & $(1.43,1.57)$ & $(7.76,7.84)$
& $(27.85,27.95)$ & $(4.83,5.01)$ & $(1.96,2.04)$ & $(4.53,4.67)$ \\
& Implicit
& $\textbf{(29.78,29.82)}$ & $\textbf{(1.94,2.06)}$ & $\textbf{(8.52,8.66)}$
& $\textbf{(29.03,29.17)}$ & $\textbf{(5.57,5.63)}$ & $\textbf{(2.01,2.07)}$ & $\textbf{(4.62,4.78)}$ \\
\cmidrule(lr){1-9}
& CoT
& $(29.97,30.03)$ & $(28.88,29.12)$ & $(40.87,41.13)$
& $(18.46,18.54)$ & $(41.26,41.34)$ & $(19.83,20.17)$ & $(30.54,30.66)$ \\
\multirow{-1}{*}{\textit{\makecell[l]{WizardLM}}}
& Explicit
& $(27.76,28.04)$ & $(34.44,34.56)$ & $\textbf{(46.18,46.22)}$
& $\textbf{(23.04,23.16)}$ & $\textbf{(45.11,45.29)}$ & $\textbf{(25.97,26.03)}$ & $\textbf{(33.07,33.13)}$ \\
& Implicit
& $\textbf{(31.74,31.86)}$ & $\textbf{(36.76,36.84)}$ & $(43.06,43.14)$
& $(21.66,21.74)$ & $(42.07,42.13)$ & $(22.71,22.89)$ & $(30.63,30.77)$ \\
\bottomrule

\end{tabular}
}
\caption{Estimated 95\% confidence intervals calculated using $t$-distribution for all main results.}
\label{tab:ci}
\end{table*}

\begin{table*}[!t]
\centering
\renewcommand\arraystretch{1.3}
\setlength{\tabcolsep}{4.85pt}
\scalebox{0.68}{
\begin{tabular}{llccccccccccccc}
\hline
\toprule
{\multirow{2.2}{*}{\textbf{Model}}}
& {\multirow{2.2}{*}{\textbf{Strategy}}} 
& \multicolumn{3}{c}{\textbf{GSM8K}} 
& \multicolumn{3}{c}{\textbf{ASDiv}}
& \multicolumn{3}{c}{\textbf{AQuA}}
& \multicolumn{4}{c}{\textbf{PRM800K}} \\
\cmidrule(lr){3-5} \cmidrule(lr){6-8} \cmidrule(lr){9-11} \cmidrule(lr){12-15}
& & \textbf{\texorpdfstring{label\textsubscript{ans}}} & \textbf{edit} & \textbf{solve} & \textbf{\texorpdfstring{label\textsubscript{ans}}} & \textbf{edit} & \textbf{solve} & \textbf{\texorpdfstring{label\textsubscript{ans}}} & \textbf{edit} & \textbf{solve} & 
\textbf{\texorpdfstring{label\textsubscript{ans}}} & \textbf{\texorpdfstring{label\textsubscript{step}}} & \textbf{edit} & \textbf{solve} \\
\midrule
& CoT+
& $\textbf{86.4}_{0.4}$ & $82.8_{0.4}$ & $92.7_{0.1}$
& $90.6_{0.5}$ & $82.2_{1.1}$ & $90.9_{0.3}$
& $\textbf{65.7}_{1.5}$ & $33.2_{1.5}$ & $56.1_{0.9}$
& $\textbf{30.2}_{1.1}$ & $\textbf{51.8}_{1.0}$ & $19.0_{1.9}$ & $43.2_{0.4}$ \\
\multirow{-2}{*}{\textit{\makecell[l]{Llama 3 70B \\ Instruct}}}
& Implicit
& $84.0_{0.7}$ & $\textbf{84.8}_{0.1}$ & $\textbf{93.3}_{0.4}$
& $\textbf{91.4}_{0.5}$ & $\textbf{84.9}_{0.7}$ & $\textbf{91.1}_{0.2}$
& $56.6_{0.2}$ & $\textbf{37.6}_{1.3}$ & $\textbf{56.4}_{0.4}$
& $19.2_{0.2}$ & $50.0_{0.6}$ & $\textbf{26.5}_{1.9}$ & $\textbf{48.4}_{0.6}$ \\
\cmidrule(lr){1-15}
& CoT+
& $50.9_{1.1}$ & $20.8_{1.2}$ & $\textbf{64.3}_{1.2}$
& $54.7_{0.2}$ & $50.8_{1.2}$ & $\textbf{77.3}_{1.1}$
& $35.6_{1.6}$ & $7.2_{1.4}$ & $\textbf{21.8}_{0.2}$
& $26.2_{0.3}$ & $38.1_{1.7}$ & $8.2_{0.7}$ & $12.1_{0.7}$ \\
\multirow{-2}{*}{\textit{\makecell[l]{Command R}}}
& Implicit
& $\textbf{64.2}_{0.8}$ & $\textbf{31.1}_{1.2}$ & $60.5_{1.2}$
& $\textbf{60.3}_{0.7}$ & $\textbf{51.4}_{0.7}$ & $70.1_{0.6}$
& $\textbf{39.8}_{1.3}$ & $\textbf{11.2}_{1.2}$ & $19.1_{0.3}$
& $\textbf{56.0}_{0.9}$ & $\textbf{43.4}_{1.2}$ & $\textbf{8.8}_{0.4}$ & $\textbf{14.8}_{0.7}$ \\
\cmidrule(lr){1-15}
& CoT+
& $69.7_{0.3}$ & $52.1_{0.7}$ & $77.2_{0.3}$
& $80.1_{0.3}$ & $63.2_{0.6}$ & $\textbf{86.3}_{0.9}$
& $46.9_{0.8}$ & $12.8_{2.0}$ & $32.8_{2.1}$
& $17.3_{0.5}$ & $34.2_{0.1}$ & $12.2_{1.3}$ & $\textbf{22.9}_{0.5}$ \\
\multirow{-2}{*}{\textit{\makecell[l]{Command R+}}}
& Implicit
& $\textbf{71.9}_{0.4}$ & $\textbf{62.0}_{0.8}$ & $\textbf{79.9}_{0.8}$
& $\textbf{82.6}_{0.2}$ & $\textbf{70.7}_{1.3}$ & $85.3_{0.3}$
& $\textbf{47.6}_{0.9}$ & $\textbf{16.8}_{1.1}$ & $\textbf{35.8}_{1.1}$
& $\textbf{59.5}_{1.2}$ & $\textbf{39.2}_{0.2}$ & $\textbf{16.6}_{0.8}$ & $21.1_{0.6}$ \\
\cmidrule(lr){1-15}
& CoT+
& $\textbf{64.2}_{0.5}$ & $49.8_{0.3}$ & $\textbf{80.5}_{1.2}$
& $50.4_{0.4}$ & $70.6_{1.2}$ & $83.6_{0.4}$
& $41.9_{0.9}$ & $14.5_{1.1}$ & $36.3_{1.3}$
& $65.0_{0.7}$ & $36.7_{0.4}$ & $11.2_{0.9}$ & $27.9_{1.0}$ \\
\multirow{-2}{*}{\textit{\makecell[l]{Command R \\ Refresh}}}
& Implicit
& $62.5_{1.0}$ & $\textbf{57.4}_{0.7}$ & $79.2_{0.7}$
& $\textbf{70.4}_{1.1}$ & $\textbf{72.2}_{0.3}$ & $\textbf{84.8}_{0.5}$
& $\textbf{50.7}_{0.9}$ & $\textbf{16.6}_{0.6}$ & $\textbf{40.5}_{0.8}$
& $\textbf{71.1}_{0.9}$ & $\textbf{53.7}_{0.8}$ & $\textbf{11.8}_{1.1}$ & $\textbf{32.1}_{0.7}$ \\
\cmidrule(lr){1-15}
& CoT+
& $45.0_{0.4}$ & $45.1_{0.2}$ & $85.2_{0.6}$
& $66.5_{1.5}$ & $78.5_{0.2}$ & $88.9_{0.1}$
& $62.0_{1.9}$ & $18.5_{0.4}$ & $45.4_{1.3}$
& $61.5_{1.2}$ & ${47.1}_{1.1}$ & $16.7_{1.3}$ & ${27.5}_{0.2}$ \\
\multirow{-2}{*}{\textit{\makecell[l]{Command R+ \\ Refresh}}}
& Implicit
& $\textbf{47.2}_{0.8}$ & $\textbf{62.8}_{0.9}$ & $\textbf{86.3}_{0.8}$
& $\textbf{79.6}_{0.2}$ & $\textbf{81.9}_{1.0}$ & $\textbf{90.4}_{0.4}$
& $\textbf{63.3}_{1.2}$ & $\textbf{21.5}_{1.1}$ & $\textbf{47.8}_{1.2}$
& $\textbf{73.9}_{0.8}$ & $\textbf{47.8}_{0.7}$ & $\textbf{18.7}_{1.1}$ & $\textbf{29.7}_{0.7}$ \\
\cmidrule(lr){1-15}
& CoT+
& $52.5_{1.0}$ & $1.2_{0.2}$ & $30.4_{1.4}$
& $46.6_{1.2}$ & $12.5_{0.2}$ & $61.1_{0.3}$
& $\textbf{57.0}_{1.7}$ & $1.2_{0.1}$ & $\textbf{11.2}_{0.3}$
& $\textbf{66.3}_{1.2}$ & $47.2_{0.5}$ & $2.8_{0.3}$ & $5.4_{0.5}$ \\
\multirow{-2}{*}{\textit{\makecell[l]{Titan Text G1 \\ Express}}}
& Implicit
& $\textbf{69.3}_{0.5}$ & $\textbf{3.0}_{0.4}$ & $\textbf{34.7}_{0.9}$
& $\textbf{60.2}_{0.4}$ & $\textbf{12.9}_{0.1}$ & $\textbf{62.8}_{0.4}$
& $49.6_{0.8}$ & $\textbf{1.4}_{0.1}$ & $11.0_{0.2}$
& $45.7_{0.9}$ & $\textbf{49.7}_{1.0}$ & $\textbf{4.0}_{0.5}$ & $\textbf{5.8}_{0.5}$ \\
\cmidrule(lr){1-15}
& CoT+
& $83.1_{0.7}$ & $69.3_{1.2}$ & $\textbf{92.0}_{0.5}$
& $94.2_{0.2}$ & $80.3_{0.2}$ & $90.4_{0.7}$
& $78.9_{0.2}$ & $42.9_{0.4}$ & $57.1_{0.3}$
& $63.2_{0.6}$ & $62.3_{0.1}$ & $33.3_{1.3}$ & $59.7_{0.2}$ \\
\multirow{-2}{*}{\textit{\makecell[l]{WizardLM}}}
& Implicit
& $\textbf{86.9}_{0.5}$ & $\textbf{82.3}_{0.9}$ & $91.5_{0.7}$
& $\textbf{95.0}_{0.2}$ & $\textbf{82.6}_{0.2}$ & $\textbf{90.6}_{0.3}$
& $\textbf{79.5}_{0.1}$ & $\textbf{44.1}_{0.5}$ & $\textbf{62.1}_{0.6}$
& $\textbf{69.0}_{0.4}$ & $\textbf{62.6}_{0.5}$ & $\textbf{35.7}_{1.3}$ & $\textbf{61.1}_{0.6}$ \\

\bottomrule
\hline
\end{tabular}
}
\caption{
Results of CoT prompting with extended context (CoT+) to match the sequence length of implicit prompting. Note that CoT+ includes further, diverse exemplars that implicit prompting does not contain. We report the accuracy of the final numerical result for all tasks except \texorpdfstring{label\textsubscript{ans} and }  \texorpdfstring{label\textsubscript{step}}, where we report the weighted F1-score of the binary label. Results are averaged over five runs to account for small variations in model-generated outputs, likely due to dynamic batching in the APIs.
}
\label{tab:analysis_result}
\end{table*}

\subsection{Answer Generation}\label{sec:ans_gen}

In Table \ref{tab:models} we show the LLMs used to generate answers to the questions in each dataset, for both the training few-shot examples and the test samples. For the training set, we only generate incorrect answers with the listed models, while all correct answers are generated with GPT-4 or extracted from the original dataset where possible. For test samples, we use these models to generate both correct and incorrect answers. We do not run this generation step for PRM800K, as this dataset already contains annotated correct and incorrect answers.

\subsection{Test Set Construction}

For GSM8K and ASDiv we use all test samples, with or without the correct/incorrect answers generated as per Section \ref{sec:ans_gen}, depending on the task. For AQuA, we make minor changes to the test set before generating the answers. PRM800K already contains CoT-style answers, though these are annotated for correctness at the intermediary reasoning step level, and not as a whole. Thus, we adapt this dataset to our tasks. We illustrate these adaptations below.

\paragraph{AQuA} contains, in its original version, multiple-choice questions associated with five answer options, only one of which is correct. In our experiments, we discard the options and prompt the model to generate open-ended answers. For ease of verifying the correctness of the answers at test time, we drop from the test set all samples where the golden answer is non-numerical.

\paragraph{PRM800K} comprises questions and the respective answers, split into intermediary reasoning steps. Each step is labelled by human annotators as correct (label 1), incorrect (label $-$1), or neutral (label~0). Some samples are associated with a series of steps that lead to the correct solution, while others contain errors that impact the final solution. For the step labelling task, we use the reasoning steps that are annotated as either correct or incorrect. We append each of them to the previous context where available, i.e., the (correct or neutral) steps that precede it in the answer. For the answer labelling and editing tasks, we join the individual steps and label the resulting answer as correct if all steps are labelled as either correct or neutral, and incorrect if at least one of the steps contains errors.

\section{Statistical Significance of Results}\label{sec:statistics}

 To confirm the statistical significance of our results, we compute the 95\% $t$-based confidence intervals of the accuracies obtained with all seven LLMs on all datasets (GSM8K, ASDiv, AQuA, PRM800K) and tasks (label\textsubscript{ans}, label\textsubscript{step}, edit, solve) with each prompting strategy (CoT, implicit learning, explicit learning). The confidence intervals are shown in Table~\ref{tab:ci}.

\section{Extended Context Length Experiments}

\subsection{CoT+}

Table~\ref{tab:analysis_result} illustrates the fine-grained results of CoT+, and compares them to implicit learning prompting. Firstly, we note that in the large majority of cases ($\sim$80\%) adding further few-shot examples to the CoT prompt results in better or similar ($<$1\% difference) performance than the same setup with fewer examples (shown in Table~\ref{tab:main_result}). In a minority of cases, however, we observe that performance declines. This is consistent with previous findings that more examples do not strictly guarantee performance improvements \cite{zhao2023incontextexemplarscluesretrieving}, especially in complex tasks \cite{opedal2024mathgapoutofdistributionevaluationproblems}. Indeed, instances where performance declines are predominantly concentrated in the PRM800K dataset, which contains particularly challenging problems. Notably, prompting for implicit learning outperforms CoT+ in over 80\% of cases. This demonstrates that the advantage of implicit learning is indeed due to the presence of incorrect answers rather than increased context length or other effects.\vspace{6pt}

\begin{table}[!t]
\centering
\renewcommand\arraystretch{1.1}
\setlength{\tabcolsep}{12pt}
\scalebox{0.86}{
\begin{tabular}{llcc}
\hline
\toprule
{\multirow{2.2}{*}{\textbf{Dataset}}}
& {\multirow{2.2}{*}{\textbf{Task}}} 
& \multicolumn{2}{c}{\textbf{Strategy}} \\
\cmidrule(lr){3-4}
& & CoT-2 & Implicit \\
\midrule
& \textbf{\texorpdfstring{label\textsubscript{ans}}}
& $\textbf{76.3}_{0.2}$ & $75.0_{0.1}$ \\
\multirow{-1}{*}{\textbf{GSM8K}}
& \textbf{edit}
& $51.6_{0.6}$ & $\textbf{63.0}_{0.9}$ \\
& \textbf{solve}
& $76.9_{1.0}$ & $\textbf{81.1}_{0.5}$ \\

\cmidrule(lr){1-4}
& \textbf{\texorpdfstring{label\textsubscript{ans}}}
& $79.5_{0.0}$ & $\textbf{84.3}_{0.1}$ \\
\multirow{-1}{*}{\textbf{ASDiv}}
& \textbf{edit}
& $66.6_{0.7}$ & $\textbf{72.0}_{0.4}$ \\
& \textbf{solve}
& $85.0_{0.3}$ & $\textbf{85.5}_{0.6}$ \\

\cmidrule(lr){1-4}
& \textbf{\texorpdfstring{label\textsubscript{ans}}}
& $69.9_{0.0}$ & $\textbf{70.7}_{0.1}$ \\ 
& \textbf{\texorpdfstring{label\textsubscript{step}}}
& $34.4_{0.0}$ & $\textbf{35.5}_{0.1}$ \\
\multirow{-2}{*}{\textbf{PRM800K}}
& \textbf{edit}
& $15.0_{1.0}$ & $\textbf{17.1}_{1.0}$ \\
& \textbf{solve}
& $25.5_{0.7}$ & $\textbf{28.2}_{1.0}$ \\

\bottomrule
\hline
\end{tabular}
}
\caption{
Results of CoT with two correct answers (CoT-2) and implicit learning. We report the accuracy of the final numerical result for all tasks except \texorpdfstring{label\textsubscript{ans} and }  \texorpdfstring{label\textsubscript{step}}, where we report the weighted F1-score of the binary label. Results are averaged over five runs.
}
\label{tab:preliminary_result}
\vspace{10pt}
\end{table}
\begin{table}[!t]
\centering

\renewcommand\arraystretch{1.2}
\setlength{\tabcolsep}{21pt}
\scalebox{0.8}{
\begin{tabular}{lcc}

\hline
\toprule
\textbf{Strategy} 
& \textbf{\makecell[l]{ROUGE-1}} 
& \textbf{\makecell[l]{ROUGE-L}}\\
\midrule
CoT & 0.18 & 0.13 \\
Implicit
& 0.26 & 0.19 \\
Explcit
& 0.19 & 0.14 \\

\bottomrule
\hline
\end{tabular}
}
\caption{
ROUGE-1 and ROUGE-L recall scores between generated and in-context rationales.}
\label{tab:rouge}
\end{table}

\subsection{CoT-2}\label{sec:cot-2}
Table~\ref{tab:preliminary_result} displays the results of running CoT-2, compared to prompting for implicit learning. We observe that implicit prompting is superior to CoT-2, with the largest overall advantage in the editing and solving tasks. Surprisingly, observing incorrect answers alongside correct ones does not help the LLM label new answers for correctness in the case of GSM8K. Overall, however, the advantage of implicit prompting over CoT-2 is consistent, further evidencing that LLMs benefit from incorrect answers in their context more than from additional correct answers.

It should be noted that this experiment---which was run early on in our work---tests one model (Command R+) on three datasets (AQuA was not yet part of our test suite). LLM instructions also have slightly different wording to those used in our main setup (hence the minor discrepancies in the final values). In particular, the labelling tasks here are set up so that the LLM outputs the label directly, as opposed to a rationale justifying its choice first, followed by the label. As detailed in Section~\ref{sec:tasks}, this was later updated to guarantee robustness.\vspace{6pt}

\section{ROUGE Analysis of Rationales}\label{sec:rouge}

In addition to the $n$-gram similarity analysis of Table~\ref{tab:rat_sim}, we assess overfitting of the generated rationales to the in-context ones via ROUGE-1 and ROUGE-L, under each prompting strategy. The former measures the overlap of individual words, while the latter measures the Longest Common Subsequence (LCS). Recall scores for both metrics are shown in Table~\ref{tab:rouge}. Consistent with the $n$-gram similarity analysis, we observe that explicit prompting leads to higher ROUGE scores, and thus generated rationales that are more similar to the exemplar ones, compared with CoT and implicit prompting.\vspace{6pt}

\begin{figure}[!b]
    \centering
    \includegraphics[width=\linewidth]{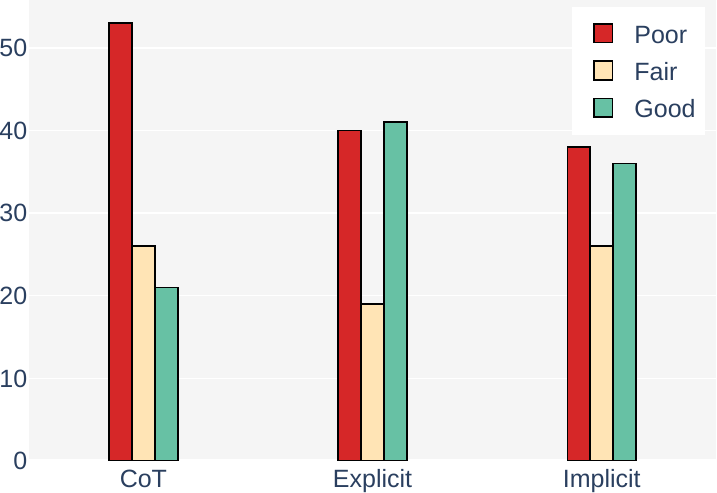}
    \caption{Fine-grained results of the human evaluation, showing the number of individual labels assigned to rationales for each prompting strategy. Explicit and implicit learning perform fairly similarly, with explicit learning obtaining a slightly higher number of labels at both extremes (`poor' and `good') and implicit learning earning more mid-range labels (`fair'). In contrast, rationales output with the aid of CoT prompting are mainly scored as `poor'.}
    \label{fig:humaneval}
\end{figure}

\section{Human Evaluation}\label{sec:human_eval}
\vspace{5pt}
\subsection{Guidelines}

We report the guidelines given to annotators for the human evaluation task. Annotators were recruited among machine learning experts.

\vspace{15pt}

\noindent{\textit{\textbf{Evaluating model feedback on math reasoning questions.}}}

\vspace{8pt}

\noindent{\textit{The attached sheet contains 100 mathematical questions and the corresponding answers given by language models. Each answer is highlighted in either green (meaning it reached the correct numerical solution) or red (meaning the numerical solution reached is wrong).}}

\noindent{\textit{For each answer, three LLMs have generated a piece of feedback explaining why the answer is wrong or correct, your task is to score each feedback as “poor”, “fair”, or “good”.}}

\noindent{\textit{In your evaluation, you should only consider the correctness of the feedback. Did the model identify the strengths and/or weaknesses of the answer correctly? In your assessment, do not take feedback length into consideration. If two pieces of feedback both identify the same key points, they should be awarded the same score, even if one is much more succinct than the other. If a piece of feedback is completely missing however (meaning the model did not generate one), you should assign the label “poor”. Also please ignore formatting and the presence of any special tokens or characters in your evaluation, only focus on the meaning. In each row, the three models are displayed in different order to avoid annotation bias. So, for example, “Model 1” in the first row may not be the same model as “Model 1” in the second row, and so on.}}

\subsection{Fine-grained Results}

In Figure~\ref{fig:humaneval}, we show a breakdown of the labels assigned by human evaluators to model-generated rationales produced with each prompting strategy. While most rationales generated via CoT are assigned the minimum score, explicit and implicit learning prompting exhibit similar trends, with explicit learning obtaining slightly more labels at both ends of the spectrum (`poor' and `good') and implicit learning receiving more mid-range labels.

\end{document}